\RequirePackage{fix-cm}
\documentclass{svjour3}                     
\smartqed  
\usepackage{graphicx}
\usepackage{url}

\usepackage{natbib}
\usepackage{verbatim}

\usepackage{graphicx} 
\usepackage{subfigure} 
\usepackage{tikz} 
\usetikzlibrary{fit,positioning}

\usepackage{chngpage} 

\usepackage{amsmath,amssymb}
\usepackage{bbm}

\usepackage[linesnumbered, ruled, procnumbered]{algorithm2e}
\usepackage{hyperref}
\hypersetup{
colorlinks,
citecolor=blue,
filecolor=blue,
linkcolor=blue,
urlcolor=blue
}

\usepackage{multirow}

\newcommand{\X}{\mathbf{X}}
\newcommand{\W}{\mathbf{W}}

\newcommand{\Y}{\mathbf{Y}}
\newcommand{\x}{\mathbf{x}}
\newcommand{\w}{\mathbf{w}}

\newcommand{\y}{\mathbf{y}}
\newcommand{\Pm}{\mathbf{P}}
\newcommand{\eps}{\boldsymbol{\epsilon}}

\newcommand{\R}{\mathbb{R}}
\newcommand{\N}{\mathcal{N}}

\newcommand{\OO}{\mathcal{O}}

\begin{document}

\title{Distributed Bayesian Matrix Factorization with Limited Communication\thanks{This is the pre-print version. The paper is published in Machine Learning journal. Definitive version DOI: 10.1007/s10994-019-05778-2.}
}


\author{Xiangju Qin$^{\ddag}$  \and
        Paul Blomstedt$^{\ddag}$ \and
        Eemeli Lepp\"{a}aho \and
        Pekka Parviainen \and
        Samuel Kaski
}
\authorrunning{Qin et al.} 

\institute{
              Helsinki Institute for Information Technology HIIT, Department of Computer Science, Aalto University \\
              \email{first.last@aalto.fi}            \\
              \emph{Present address of P. Parviainen:} Department of Informatics, University of Bergen\\
{\small$^{\ddag}$~These authors contributed equally to this work.}
}

\date{Received: date / Accepted: date}

\maketitle

\begin{abstract}
Bayesian matrix factorization (BMF) is a powerful tool for producing low-rank representations of matrices and for predicting missing values and providing confidence intervals. Scaling up the posterior inference for massive-scale matrices is challenging and requires distributing both data and computation over many workers, making communication the main computational bottleneck. Embarrassingly parallel inference would remove the communication needed, by using completely independent computations on different data subsets, but it suffers from the inherent unidentifiability of BMF solutions. We introduce a hierarchical decomposition of the joint posterior distribution, which couples the subset inferences, allowing for embarrassingly parallel computations in a sequence of at most three stages. Using an efficient approximate implementation, we show improvements empirically on both real and simulated data. Our distributed approach is able to achieve a speed-up of almost an order of magnitude over the full posterior, with a negligible effect on predictive accuracy. Our method outperforms state-of-the-art embarrassingly parallel MCMC methods in accuracy, and achieves results competitive to other available distributed and parallel implementations of BMF.
\end{abstract}

\section{Introduction}
\label{intro}
Latent variable models based on matrix factorization have in recent years become one of the most popular and successful approaches for matrix completion tasks, such as collaborative filtering in recommender systems \citep{Koren+others:2009} and drug discovery \citep{Cobanoglu+others:2013}. The main idea in such models is, given a matrix of observed values $\Y\in \R^{N\times D}$, to find two matrices $\X\in \R^{N\times K}$ and $\W\in \R^{D\times K}$ with $K\ll N,D$, such that their product forms a low-rank approximation of $\Y$: 
\begin{equation}\label{eq:MF}
\Y \approx \X \W^\top.
\end{equation}
In matrix completion, the matrix $\Y$ is typically very sparsely observed, and the goal is to predict unobserved matrix elements based on the observed ones. 

A standard way of dealing with high levels of unobserved elements in matrix factorization is to model the observed values only (instead of imputing the missing values), using regularization to avoid overfitting. A probabilistically justified regularized matrix factorization model was first introduced by \citet{Salakhutdinov+Mnih:2008a}, and subsequently extended to a fully Bayesian formulation (called Bayesian Probabilistic Matrix Factorization, BPMF) \citep{Salakhutdinov+Mnih:2008b}. This formulation sidesteps the difficulty of choosing appropriate values for the regularization parameters by considering them as hyperparameters, placing a hyperprior over them and using Markov chain Monte Carlo (MCMC) to perform posterior inference. Additional advantages of the fully Bayesian approach include improved predictive accuracy, quantification of the uncertainty in predictions, the ability to incorporate prior knowledge, as well as flexible utilization of side-information \citep{Adams+others:2010,Park+others:2013,Porteous+others:2010,Simm+others:2015}. 

Given the appeal and many advantages of Bayesian matrix factorization, applying it also to massive-scale matrices would be attractive but scaling up the posterior inference has proven difficult, and calls for distributing both data and computation over many workers. So far only very few distributed implementations of BMF have been presented in the literature. Recently, \citet{Ahn+others:2015} proposed a solution based on distributed stochastic gradient Langevin dynamics (DSGLD), and showed empirically that BMF with DSGLD achieves the same level of predictive performance as Gibbs sampling. However, the convergence efficiency of the DSGLD solution is constrained by several factors such as the nee for careful tuning of the learning rate $\epsilon_{t}$ and for using an orthogonal group partition\footnote{This partition is used to avoid conflicting access to parameters among parallel workers, and refers to a partition scheme in which rows and columns included in one block do not appear in the other ones.} for training. When a model is trained with blocks in an orthogonal group, in each iteration it only makes use of a small subset of the full data set for learning, which could lead to an estimate with higher variance and slowing down the convergence speed. \cite{Simsekli+others:2015} developed a similar distributed MCMC method based on SGLD for large generalised matrix factorization problems, which they called Parallel SGLD (PSGLD); see also \citet{Simsekli+others:2017} for an application of the method to non-negative matrix factorization (NMF). Different from DSGLD, PSGLD is implemented such that for each iteration, only blocks of $\W$ instead of the whole $\W$ need to be transferred among parallel workers. Nevertheless, this solution suffers from the same issues as DSGLD. 
\citet{VanderAa+others:2017} presented a distributed high-performance implementation of BPMF with Gibbs sampling using the TBB and GASPI libraries, and provided an empirical comparison with other state-of-the-art distributed high-performance parallel implementations. 
They found that a significant speed-up could only be achieved with a limited number of workers, after which the addition of more workers eventually leads to a dramatic drop in parallel computation efficiency due to the increased communication overhead \citep{vander2016distributed, VanderAa+others:2017}. 
Therefore, a key factor in devising even more scalable distributed solutions is to be able to minimize communication between worker nodes.

One of the most promising directions in large-scale Bayesian computation in recent years has been \emph{embarrassingly parallel} MCMC, a family of essentially communication-free algorithms, where the data are first partitioned into multiple subsets and independent sampling algorithms are then run on each subset in parallel \citep{Minsker+others:2014,Neiswanger+others:2014,Scott+others:2016,Srivastava+others:2015,Wang+Dunson:2013,Wang+others:2014,Wang+others:2015}. In these algorithms, communication only takes place in a final aggregation step, which combines the subset posteriors to form an approximation to the full-data posterior. A key factor which limits the applicability of such methods for BMF models is that Equation~(\ref{eq:MF}) can be solved only up to orthogonal transformations. 
Each subset posterior can therefore converge to any of an infinite number of modes, making the aggregation step difficult to carry out in a meaningful way.
Previous embarrassingly parallel MCMC algorithms have only been applied in cases where the model is unidentified up to a finite number of solutions \citep[e.g.][]{Nemeth+Sherlock:2017} but are not applicable in a continuum of unidentifiable cases.

In this paper, we introduce an approach which addresses the unidentifiability issue by introducing dependencies between the subset posterior inferences, while limiting the communication between workers. We will draw inspiration from the observation that even though an infinite number of solutions to Equation~(\ref{eq:MF}) exist in principle, in practical computation with a finite number of observations, a sampler with finite chain-length will only explore a small number of solutions, each corresponding to a separate mode. 
The key idea is to encourage the samplers in all subsets to target the same set of solutions; note that this does not restrict generality as the standard way of finding other modes, by employing additional chains, is available here as well.

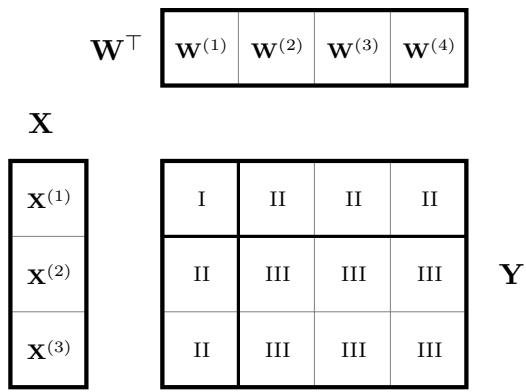
\begin{figure}
	\begin{center}
		\begin{tikzpicture}
    		\draw[help lines] (2,0) grid (6,3);
    		\node[fit={(2,0) (6,3)}, inner sep=0pt, draw=black, ultra thick] (Y) {};
    		\node[fit={(2,0) (3,3)}, inner sep=0pt, draw=black, very thick] (Yi1) {};
    		\node[fit={(2,2) (6,3)}, inner sep=0pt, draw=black, very thick] (Y1j) {};
		\node[align=center] at (2.5+0.02,2.5) {I};
	    \node[align=center] at (3.5+0.02,2.5) {II};
		\node[align=center] at (4.5+0.02,2.5) {II};
	    \node[align=center] at (5.5+0.02,2.5) {II};
		\node[align=center] at (2.5+0.02,1.5) {II};
	    \node[align=center] at (3.5+0.02,1.5) {III};
		\node[align=center] at (4.5+0.02,1.5) {III};
	    \node[align=center] at (5.5+0.02,1.5) {III};
	    	\node[align=center] at (2.5+0.02,0.5) {II};
	    \node[align=center] at (3.5+0.02,0.5) {III};
		\node[align=center] at (4.5+0.02,0.5) {III};
	    \node[align=center] at (5.5+0.02,0.5) {III};
    		\draw[help lines] (0,0) grid (1,3);
    		\node[fit={(0,0) (1,3)}, inner sep=0pt, draw=black, ultra thick] (X) {};
	    \node[align=center,font=\small] at (0.5+0.02,2.5) {$\X^{(1)}$};
	    \node[align=center,font=\small] at (0.5+0.02,1.5) {$\X^{(2)}$};
	    	\node[align=center,font=\small] at (0.5+0.02,0.5) {$\X^{(3)}$};
    		\draw[help lines] (2,4) grid (6,5);
    		\node[fit={(2,4) (6,5)}, inner sep=0pt, draw=black, ultra thick] (W) {};
	    \node[align=center,font=\small] at (2.5+0.02,4.5) {$\W^{(1)}$};
	    \node[align=center,font=\small] at (3.5+0.02,4.5) {$\W^{(2)}$};
	    	\node[align=center,font=\small] at (4.5+0.02,4.5) {$\W^{(3)}$};
	    \node[align=center,font=\small] at (5.5+0.02,4.5) {$\W^{(4)}$};
    	    \node[align=center,font=\large] at (6.5+0.1,1.5) {$\Y$};
    	    \node[align=center,font=\large] at (0.5-0.1,3.5) {$\X$};
    	    \node[align=center,font=\large] at (1.5-0.1,4.5) {$\W^\top$};
		\end{tikzpicture}
	\end{center}
	\caption{An example illustrating posterior propagation (PP) for a data matrix $\Y$ partitioned into $3 \times 4$ subsets. Subset inferences proceed in three successive stages, with posteriors obtained in one stage being propagated as priors to the next one; the numbers in the $\Y$ matrix denote the stage in which the particular subset is processed. Within each stage, the subsets are processed in parallel with no communication.}\label{fig:PP}
\end{figure}

In large BMF problems we partition the data matrix $\Y$ along both rows and columns, effectively making different subsets dependent on parameters shared by subsets on the same rows and columns. To implement the dependencies in the inference, we divide the subsets into three groups, 
which are processed in a hierarchy of three consecutive stages (see Figure~\ref{fig:PP}). The posterior distributions obtained in each stage are propagated forwards and used as priors in the following stage. This way, communication only takes place between the stages and not between the subsets within a stage, and each subset inference is regularized using information from the relevant subset inference in the preceding stage.
Note that within each stage, we perform the inference for subsets in parallel. Thus, for a partition scheme with $r \times c$ subsets, the maximum number of parallel workers that can be used by our algorithm is equal to the number of subsets in the third stage, i.e. $(r-1)\times(c-1)$.
We refer to the proposed procedure as \emph{posterior propagation (PP)}. Table \ref{Tab:comparison_communication_cost} compares the computational characteristics of PP with previous approaches for parallel and distributed BMF / NMF.

\begin{table}[!htb]
\caption{Comparison of different implementations of parallel and distributed BMF / NMF for a data matrix $\Y \in \R^{N \times D}$. 
We assume a a maximum of $U$ compute nodes available for distributed inference and a partition scheme of $(\sqrt{U} + 1) \times (\sqrt{U} + 1)$ for all methods. The number of latent dimensions and iterations are $K$ and $T$, respectively.
For simplicity, $\Y$ is assumed to be fully observed. For our method, the $K^{2}$ term is due to using a full covariance matrix for the posterior of parameters $\W$ and $\X$. For BMF + DSGLD, the $\sqrt{U} + 1$ term is applied only when $\sqrt{U}+1$ parallel chains are deployed. Note that communication cost is incurred only when the model is running on multiple compute nodes (distributed memory). Also note that the cost of loading the data is ignored; there is no communication incurred during inference for the embarrassingly parallel methods.}
\label{Tab:comparison_communication_cost}
\renewcommand{\arraystretch}{1.35}
\begin{adjustwidth}{-.15in}{-.15in}
\begin{center}
{\fontsize{7.25pt}{7.25pt}\selectfont
\begin{tabular}{l|c|c|c|c}
  \hline

\multirow{3}{*}{Models} & Tune learning & Shuffle & Optimal & \multirow{3}{*}{Communication cost}\\ 
 & rate or & train data & partition &  \\ 
 & hyperparam. & per iter. &  &  \\ \hline
 
BMF + DSGLD & \multirow{2}{*}{\checkmark} & \multirow{2}{*}{\checkmark} & Squared & \multirow{2}{*}{$\OO( (N+D) K (\sqrt{U}+1) T )$} \\ 
\citep{Ahn+others:2015} &  &  & orthogonal &   \\ \hline

NMF + PSGLD & \multirow{2}{*}{\checkmark} & \multirow{2}{*}{\checkmark} & Squared & \multirow{2}{*}{$\OO(D K T)$} \\ 
\citep{Simsekli+others:2017} &  &  & orthogonal & \\ \hline

Distributed BPMF & \multirow{2}{*}{-} & \multirow{2}{*}{-} & \multirow{2}{*}{Load balance} & \multirow{2}{*}{$\OO( (N + D) K (U-1) T )$} \\ 
\citep{VanderAa+others:2017} &  &  &  &   \\ \hline

Proposed method & - & - & Flexible & $\OO( (N + D) (K+K^{2}) \sqrt{U})$ \\ 

\hline
\end{tabular}
}
\end{center}
\end{adjustwidth}
\end{table}

\subsection{Contributions and overview of the paper}

The main contributions of our paper are as follows: %
In Section~\ref{sec:decomposition}, we introduce a hierarchical, exact decomposition of the joint posterior of the BMF model parameters, which makes possible embarrassingly parallel computations over data subsets in a sequence of at most three stages, limiting all communication to take place between the stages. 
This decomposition is computationally intractable in general; however, in Section~\ref{sec:approx_inference} we build on it to develop a MCMC-based approximate inference scheme for BMF. In the numerical experiments of Section~\ref{sec:experiments}, we show empirically, with both real and simulated data, that the proposed distributed approach is able to achieve a speed-up of almost an order of magnitude over the full posterior, with a negligible effect on predictive accuracy, compared to MCMC inference on the full data. In the experiments, the method also significantly outperforms state-of-the-art embarrassingly parallel MCMC methods in accuracy, and achieves competitive results compared to other available distributed and parallel implementations of BMF.

\section{Background}

\subsection{Bayesian matrix factorization}\label{sec:BMF}

Let $\Y\in \R^{N\times D}$ be a partially observed data matrix, $\X \in \R^{N\times K}= (\x_1,\ldots,\x_N)^\top$ and $\W = (\w_1,\ldots,\w_D)^\top\in \R^{D\times K}$ be matrices of unknown parameters. The general Bayesian matrix factorization (BMF) model is then specified by the likelihood
\begin{equation}\label{eq:BMF_likelihood}
p(\Y|\X,\W) = \prod_{n=1}^N \prod_{d=1}^D \left[p\left(y_{nd}|\x_n^\top\w_d\right) \right]^{\boldsymbol{1}_{nd}},
\end{equation}
which is a probabilistic version of Equation~(\ref{eq:MF}). Here $\boldsymbol{1}_{nd}$ denotes an indicator function which equals 1 if the element $y_{nd}$ is observed and 0 otherwise. 

While the general BMF model is agnostic to the choice of distributional form, in many applications, the elements $y_{nd}$ of the data matrix are assumed to be normally distributed, conditionally on the parameter vectors $\x_n$ and $\w_d$,
\begin{equation}\label{eq:BMF_likelihood_normal}
p\left(y_{nd}|\x_n^\top\w_d\right) = \mathcal{N}\left(y_{nd}|\x_n^\top\w_d,\tau^{-1}\right),
\end{equation}
where $\tau$ denotes the noise precision. Note that some formulations specify an individual precision $\tau_d$ for each column. To complete the Bayesian model, priors are placed on the model parameters $\X$ and $\W$, commonly normal priors specified as  
\begin{subequations}
\begin{align}
p\left(\X|\mu_{\X},\Lambda_{\X}^{-1}\right) &= \prod_{n=1}^N \mathcal{N}_K\left(\x_n|\mu_{\X},\Lambda_{\X}^{-1}\right),\label{eq:prior_X}\\
p\left(\W|\mu_{\W},\Lambda_{\W}^{-1}\right) &= \prod_{d=1}^D \mathcal{N}_K\left(\w_d|\mu_{\W},\Lambda_{\W}^{-1}\right),\label{eq:prior_W}
\end{align}
\end{subequations}
where $\mathcal{N}_K$ denotes a $K$-dimensional normal distribution with covariance specified in terms of the precision matrix. 
The model formulation may additionally include priors on some or all of the hyperparameters $\mu_{\X},\Lambda_{\X},\mu_{\W},\Lambda_{\W}$, as well as on the data precision parameters $\tau_d$ \citep[e.g.][]{Bhattacharya+Dunson:2011,Salakhutdinov+Mnih:2008b}. 
For concreteness, we proceed in this paper with the Gaussian case, as specified in Equations~(\ref{eq:BMF_likelihood_normal}--\ref{eq:prior_W}), but note that the developments of Section~\ref{sec:decomposition}, along with Algorithm~\ref{alg:BMFPP}, are general with no reference to this choice of distributions for likelihood and priors.

\subsubsection{Unidentifiability of matrix factorization}

It is commonly known that the solution of Equation~(\ref{eq:MF}) is unique only up to orthogonal transformations. To demonstrate this unidentifiability, let $\Pm$ be any semi-orthogonal matrix for which $\Pm\Pm^\top = I_K$, $I_K$ being a $K\times K$ unit matrix, and denote $\hat{\Y}:=\X\W^\top$. Then, performing an orthogonal transformation by right-multiplying both $\X$ and $\W$ by $\Pm$, leads to
\[
\X \Pm (\W \Pm)^\top = \X \Pm\Pm^\top \W^\top = \X\W^\top = \hat{\Y},
\]
by which an uncountable number of equally good solutions to Equation~(\ref{eq:MF}) can be produced.

As a special case, let $\Pm$ be a $K\times K$ unit matrix with the $k$th diagonal element set to $-1$. The matrix $\Pm$ is then clearly orthogonal (also semi-orthogonal), since $\Pm^\top \Pm = \Pm\Pm^\top=I_K$. Right-multiplying $\X$ and $\W$ by $\Pm$ has the effect of flipping the signs of all elements in the $k$th columns of these matrices. It can then easily be verified that the product of the resulting matrices remains unchanged. Since any of the K columns of  $\X$ and $\W$ can have their signs flipped without affecting the product $\hat{\Y}$, within this family of transformations we have $2^K$ equally good solutions for Equation~(\ref{eq:MF}). 

Although the unidentifiability of a single matrix factorization task could be addressed by e.g. constraining $\W$ to be a lower triangular matrix \citep{Lopes+West:2004}, expensive communication would be needed for distributed inference schemes to ensure all the posteriors are jointly identifiable.

\subsection{Embarrassingly parallel MCMC}

Consider now a parametric model $p(\Y|\theta)$ with exchangeable observations $\Y = \{\y_1,\ldots,\y_N\}$ and parameter $\theta$ for which we wish to perform posterior inference using MCMC. If $N$ is very large, the inference may be computationally too expensive to be carried out on a single machine.
Embarrassingly parallel MCMC strategies aim to overcome this by partitioning the data $\Y$ into multiple disjoint subsets $\Y^{(1)}\cup \cdots \cup \Y^{(J)} = \Y$, and running independent sampling algorithms for each subset using a down-weighted prior $p(\theta)^{1/J}$. In most embarrassingly parallel MCMC algorithms, the aggregation of the obtained subset posteriors into a full-data posterior is based, in one way or another, on the following \emph{product density equation}:
\begin{equation}\label{eq:PDE}
p(\theta|\Y) \propto p(\theta)p(\Y|\theta) = \prod_{j=1}^J p(\theta)^{1/J}p\left(\Y^{(j)}|\theta\right),
\end{equation}
where each factor $p(\theta)^{1/J}p\left(\Y^{(j)}|\theta\right)$ constitutes an independent inference task. Aggregating the joint \emph{product density equation} with satisfactory efficiency and accuracy is in general a challenging problem, since the involved densities are unknown and represented in terms of samples. For a recent overview of various subset posterior aggregation techniques, see \citet{Angelino+others:2016}.

Standard embarrassingly parallel inference techniques relying on Equation~(\ref{eq:PDE}) are not well-suited for unidentifiable models, such as the BMF model presented in Section~\ref{sec:BMF}. 
To illustrate this, consider the following simple example: 
\begin{example}
Assume that we have observed a data matrix
\[
\Y = 
\begin{bmatrix}
1 & 4 & 16
\end{bmatrix}, 
\]
conditional on which we wish to estimate the parameters $\X$ and $\W$ of the corresponding BMF model. 
A plausible inference may then result in a bimodal posterior with high density regions around, say, the exact solutions
\[
\X=
\begin{bmatrix}
4
\end{bmatrix},
\quad \W = 
\begin{bmatrix}
0.25 & 1 & 4
\end{bmatrix}, 
\]
and $-\X,-\W$. Next, assume that we split the data into three subsets
\[
\Y^{(1)} = 
\begin{bmatrix}
1
\end{bmatrix}, \quad 
\Y^{(2)} = 
\begin{bmatrix}
4
\end{bmatrix}, \quad 
\Y^{(3)} = 
\begin{bmatrix}
16
\end{bmatrix},
\]
and conduct independent inference over each of them. Again, plausible subset inferences may accumulate posterior mass around \emph{some} set of exact solutions, say,
\begin{align*}
&\X=
\begin{bmatrix}
1
\end{bmatrix},
\quad \W^{(1)} = 
\begin{bmatrix}
1
\end{bmatrix}, \\
&\X=
\begin{bmatrix}
2
\end{bmatrix},
\quad \W^{(2)} = 
\begin{bmatrix}
2
\end{bmatrix}, \\
&\X=
\begin{bmatrix}
4
\end{bmatrix},
\quad \W^{(3)} = 
\begin{bmatrix}
4
\end{bmatrix},
\end{align*}
along with their corresponding negative solutions.  However, aggregating these inferences using Equation~(\ref{eq:PDE}) does not necessarily lead to a posterior with high density around any correct solution. 
\end{example}
Ideally, we would like all subset inferences in the above example to target the same solutions in order for them to reinforce each other. 
To do so, it is clearly necessary to impose some constraints or regularization on them. 
One way of doing this is to equip the inferences with strong enough prior information. We will build on this idea in the following section.

\section{Hierarchical Parallelization of BMF}\label{sec:decomposition}

Let us now assume that a data matrix $\Y$ has been partitioned with respect to both rows and columns into $I\times J$ subsets $\Y^{(i,j)}$, $i=1,\ldots,I$, $j = 1,\ldots, J$. It then follows from Equations (\ref{eq:BMF_likelihood}, \ref{eq:prior_X}, \ref{eq:prior_W}), that the joint posterior density of the BMF parameter matrices $\X$ and $\W$, given the partitioned data matrix $\Y$, can be factorized as 
\begin{align}
p(\X,\W|\Y) &  \propto  p(\X)\,p(\W)\,p(\Y|\X,\W) \label{eq:joint_MF_posterior}\\
&  = \prod_{i=1}^{I} p\left(\X^{(i)}\right) \prod_{j=1}^{J} p\left(\W^{(j)}\right) \prod_{i=1}^{I}\prod_{j=1}^{J} p\left(\Y^{(i,j)}|\X^{(i)},\W^{(j)}\right).\nonumber
\end{align} 
Our goal is to develop an equivalent decomposition that fulfils the apparently contradictory aims of both allowing for embarrassingly parallel computations and making the subset inferences dependent. 

\subsection{From sequential to parallel inference}

We begin with the simple case of having only three subsets. With $I=1$ and $J=3$, the parameters of the partitioned BMF model are $\X$, $\W^{(1)}$, $\W^{(2)}$ and $\W^{(3)}$. 
Sequential inference (exploiting no parallelism) over $\Y$ can then be performed in three successive stages as follows. In the first stage, the posteriors for the parameters $\X$ and $\W^{(1)}$, given $\Y^{(1)}$, are computed as
\begin{equation}\label{eq:sequential_first_stage}
p\left(\X,\W^{(1)}|\Y^{(1)}\right) \propto p(\X)p\left(\W^{(1)}\right)p\left(\Y^{(1)}|\X,\W^{(1)}\right).
\end{equation}
In the second stage, the posterior from the first stage is used as a prior for the shared parameter $\X$ to compute
\begin{align}\label{eq:sequential_second_stage}
p\left(\X,\W^{(1)},\W^{(2)}|\Y^{(1)},\Y^{(2)}\right) \propto 
&\;p\left(\X,\W^{(1)}|\Y^{(1)}\right)p\left(\W^{(2)}\right)\\
&\times p\left(\Y^{(2)}|\X,\W^{(2)}\right).\nonumber
\end{align}
In the above stage, using the posterior obtained in Equation~(\ref{eq:sequential_first_stage}) as a prior can be interpreted as a form of regularization, which encourages the inference to target the same set of modes as the first stage. Finally, using the posterior from the second stage as a prior in the third stage then gives the full-data posterior as  
\begin{align*}
p\left(\X,\W|\Y\right) &= p\left(\X,\W^{(1)},\W^{(2)},\W^{(3)}|\Y^{(1)},\Y^{(2)},\Y^{(3)}\right)\\ 
& \propto p\left(\X,\W^{(1)},\W^{(2)}|\Y^{(1)},\Y^{(2)}\right) p\left(\W^{(3)}\right)\\
& \phantom{\propto} \times p\left(\Y^{(3)}|\X,\W^{(3)}\right). 
\end{align*}
In general, a data set partitioned into $J$ subsets will require $J$ stages of sequential inference to obtain the full posterior.

We will now consider an alternative, partly parallelizable inference scheme, which begins with an initial stage identical to that of the above sequential scheme. However,  instead of processing the subsets $\Y^{(2)}$ and $\Y^{(3)}$ in sequence, we process them in parallel. Regularizing the inferences with a common informative prior, obtained in the first stage, introduces dependence between them and encourages the targeted solutions to agree with each other.   
This leads to the following decomposition:
\begin{subequations}
\begin{align}
p\left(\X,\W|\Y\right)
\propto &\;p\left(\X,\W^{(1)}|\Y^{(1)}\right)\label{eq:simple_stage_1}\\
 &\times \left[p\left(\X,\W^{(1)}|\Y^{(1)}\right)\,p\left(\W^{(2)}\right)\,
p\left(\Y^{(2)}|\X,\W^{(2)}\right)\right] \label{eq:simple_stage_2_1}\\
 &\times \left[p\left(\X,\W^{(1)}|\Y^{(1)}\right)\,p\left(\W^{(3)}\right)\,
p\left(\Y^{(3)}|\X,\W^{(3)}\right)\right]\label{eq:simple_stage_2_2}\\
&\times  p\left(\X,\W^{(1)}|\Y^{(1)}\right)^{-2},\label{eq:stage_simple_aggreg}
\end{align}
\end{subequations}
where the right-hand side of line~(\ref{eq:simple_stage_1}) corresponds to the first stage, lines (\ref{eq:simple_stage_2_1})--(\ref{eq:simple_stage_2_2}) correspond to the second stage, with the two remaining subsets now being processed in parallel.
Finally, an aggregation stage combines all of (\ref{eq:simple_stage_1})--(\ref{eq:stage_simple_aggreg}).

With $J=2$, the number of stages for the parallel scheme is exactly the same as for the sequential one. However, while the sequential scheme always requires $J$ stages, the number of stages for the parallel scheme remains constant for all $J\geq 3$. A key challenge is then to be able to carry out the aggregation stage efficiently. Strategies for aggregation are discussed further in Section~\ref{sec:aggregation}. 

\subsection{Posterior propagation}\label{sec:pp}

We will now extend the idea introduced above for arbitrary partitions of $\Y$ and show that this yields an exact decomposition of the full joint distribution (\ref{eq:joint_MF_posterior}). As $\Y$ is partitioned along both columns and rows, our hierarchical strategy is conducted in three successive stages. Communication is only required to propagate posteriors from one stage to the next, while within each stage, the subsets are processed in an embarrassingly parallel manner with no communication. The approach, coined \emph{posterior propagation} (PP), proceeds as follows:

{\bf Inference stage I.} 
Inference is conducted for the parameters of subset $\Y^{(1,1)}$:
\begin{align}
&p\left(\X^{(1)},\W^{(1)}|\Y^{(1,1)}\right) \propto \label{eq:stage1} p\left(\X^{(1)}\right)p\left(\W^{(1)}\right) p\left(\Y^{(1,1)}|\X^{(1)},\W^{(1)}\right).
\end{align}

{\bf Inference stage II.} 
Inference is conducted in parallel for parameters of subsets which share columns or rows with $\Y^{(1,1)}$. Posterior marginals from stage 1 are used as priors for the shared parameters:
\begin{subequations}
\begin{align}
&p\left(\X^{(i)},\W^{(1)}|\Y^{(1,1)},\Y^{(i,1)}\right)\label{eq:stage2a}\\ 
& \propto  p\left(\W^{(1)}|\Y^{(1,1)}\right) p\left(\X^{(i)}\right) p\left(\Y^{(i,1)}|\X^{(i)},\W^{(1)}\right),\nonumber \\
&\nonumber\\
&p\left(\X^{(1)},\W^{(j)}|\Y^{(1,1)},\Y^{(1,j)}\right)\label{eq:stage2b}\\ 
& \propto  p\left(\X^{(1)}|\Y^{(1,1)}\right) p\left(\W^{(j)}\right) p\left(\Y^{(1,j)}|\X^{(1)},\W^{(j)}\right),\nonumber 
\end{align}
\end{subequations}
for $i=2,\ldots,I$ and $j=2,\ldots,J$.

{\bf Inference stage III.} 
The remaining subsets are processed in parallel using posterior marginals propagated from stage II as priors:
\begin{align}
&p\left(\X^{(i)},\W^{(j)}|\Y^{(1,1)},\Y^{(i,1)},\Y^{(1,j)},\Y^{(i,j)}\right)  \label{eq:stage3}\\
&\propto p\left(\X^{(i)}|\Y^{(1,1)},\Y^{(i,1)}\right) p\left(\W^{(j)}|\Y^{(1,1)},\Y^{(1,j)}\right) 
p\left(\Y^{(i,j)}|\X^{(i)},\W^{(j)}\right).\nonumber 
\end{align}

{\bf Product density equation.} 
Combining the submodels in Equations~(\ref{eq:stage1}--\ref{eq:stage3}), for all $i$ and $j$, and dividing away the multiply-counted propagated posterior marginals yields the following \emph{product density equation}:
\begin{align}
&p(\X,\W|\Y) \propto \label{eq:PP_pde}\\
&p\left(\X^{(1)},\W^{(1)}|\Y^{(1,1)}\right) \nonumber\\
& \times\prod_{i=2}^{I} \left[p\left(\X^{(i)},\W^{(1)}|\Y^{(1,1)},\Y^{(i,1)}\right) p\left(\W^{(1)}|\Y^{(1,1)}\right)^{-1}\right] \nonumber\\ 
&  \times\prod_{j=2}^{J} \left[p\left(\X^{(1)},\W^{(j)}|\Y^{(1,1)},\Y^{(1,j)}\right)p\left(\X^{(1)}|\Y^{(1,1)}\right)^{-1}\right] \nonumber\\
& \times \prod_{i=2}^{I}\prod_{j=2}^{J} \Bigg[p\left(\X^{(i)},\W^{(j)}|\Y^{(1,1)},\Y^{(i,1)},\Y^{(1,j)},\Y^{(i,j)}\right) \nonumber\\
&  \phantom{\times \prod_{i=2}^{I}\prod_{j=2}^{J} \Bigg[} 
 \times p\left(\X^{(i)}|\Y^{(1,1)},\Y^{(i,1)}\right)^{-1} p\left(\W^{(j)}|\Y^{(1,1)},\Y^{(1,j)}\right)^{-1}\Bigg]\nonumber .
\end{align}
The following theorem higlights the fact that this is indeed a proper decomposition of the full posterior density.  
\begin{theorem}
Equation~(\ref{eq:PP_pde}) is, up to proportion, an exact decomposition of the full posterior $p(\X,\W|\Y)$ given in Equation~(\ref{eq:joint_MF_posterior}).
\end{theorem}
The proof of the theorem is given in Appendix \ref{sc:proof_theorem1}.

\section{Approximate inference}\label{sec:approx_inference}

In the previous section, we introduced a hierarchical decomposition of the joint posterior distribution of the BMF model, which couples inferences over subsets but allows for embarrassingly parallel computations in a sequence of (at most) three stages.  
The challenge with implementing this scheme in practice is threefold, and relates to the analytically intractable form of the BMF posterior: i) propagating posteriors efficiently from one stage to the next, ii) utilizing the posteriors of one stage as priors in the next stage, and iii) aggregating all subset posteriors at the end. 
In this section, we propose to resolve these challenges by using parametric approximations computed from subset posterior samples obtained by MCMC in each stage.
%

Computational schemes for distributed data settings, combining MCMC with propagation of information through parametric approximations have recently been explored by \citet{Xu+others:2014} and \citet{Vehtari+others:2018}. 
Nevertheless, their expectation propagation algorithms for distributed data require frequent communication among parallel workers to share global information, which could become a bottleneck for large-scale computational problems where the number of model parameters scales linearly with the number of data samples. On the other hand, the proposed method only requires communication between stages of inference. 
While our focus here is on sampling-based inference, it is worth emphasizing that the decomposition introduced in Section~\ref{sec:pp} is itself not tied to any particular inference algorithm. Thus, it could also be combined e.g. with variational inference.

\subsection{Parametric approximations for propagation of posteriors}\label{sec:parametric_approx}

We present here three alternative approaches for finding tractable approximations from posterior samples.
A generic algorithm for the proposed inference scheme using these approximations is given in {Algorithm~\ref{alg:BMFPP}}. 
%

\begin{algorithm}[!htb]
\caption{Distributed BMF with approximate posterior propagation.}\label{alg:BMFPP}

\tcp{Inference stage I}
Draw samples from posterior of BMF with data $\Y^{(1,1)}$ and priors $p(\X)$, $p(\W)$\;\label{step:infer1}
Find approximations $\widehat{p}\left(\X^{(1)}|\Y^{(1,1)}\right)$, $\widehat{p}\left(\W^{(1)}|\Y^{(1,1)}\right)$\;\label{step:approx1}

\tcp{Inference stage II}
\For {$i = 2:I$ \textrm{\textbf{in parallel}}}{
	Draw samples from posterior of BMF with data $\Y^{(i,1)}$ and priors $p(\X)$, $\widehat{p}\left(\W^{(1)}|\Y^{(1,1)}\right)$\;\label{step:infer2a}
	Find approximations $\widehat{p}\left(\X^{(i)}|\Y^{(1,1)}, \Y^{(i,1)}\right)$, $\widehat{p}\left(\W^{(1)}|\Y^{(1,1)}, \Y^{(i,1)}\right)$\;\label{step:approx2a}
}

\For {$j = 2:J$ \textrm{\textbf{in parallel}}}{
	Draw samples from posterior of BMF with data $\Y^{(1,j)}$ and priors $\widehat{p}\left(\X^{(1)}|\Y^{(1,1)}\right)$, $p(\W)$\;\label{step:infer2b}
	Find approximations $\widehat{p}\left(\X^{(1)}|\Y^{(1,1)}, \Y^{(1,j)}\right)$, $\widehat{p}\left(\W^{(j)}|\Y^{(1,1)}, \Y^{(1,j)}\right)$\;\label{step:approx2b}
}

\tcp{Inference stage III}
\For {$i = 2:I$ \textrm{\textbf{in parallel}}}{
	\For {$j = 2:J$ \textrm{\textbf{in parallel}}}{
		Draw samples from posterior of BMF with data $\Y^{(i,j)}$ and priors $\widehat{p}\left(\X^{(i)}|\Y^{(1,1)}, \Y^{(i,1)}\right)$, $\widehat{p}\left(\W^{(j)}|\Y^{(1,1)}, \Y^{(1,j)}\right)$\;\label{step:infer3}
		Find approximations $\widehat{p}\left(\X^{(i)}|\Y^{(1,1)}, \Y^{(i,1)}, \Y^{(i,j)}\right)$, $\widehat{p}\left(\W^{(j)}|\Y^{(1,1)}, \Y^{(1,j)}, \Y^{(i,j)}\right)$\;\label{step:approx3}
	}
}
\tcp{Aggregation}
Aggregate subset posteriors to approximate the full-data posterior using {\bf Algorithm \ref{alg:corrected_aggregation}}\;
\end{algorithm}

{\bf Gaussian mixture model approximation}. 
For the first approach, we note that the posterior distributions represented by the samples are typically multimodal due to the inherent unidentifiability of the BMF model. \emph{Gaussian mixture models} (GMM) are universal approximators of probability distributions, that is, given sufficiently many components, they can approximate any continuous distribution with arbitrary accuracy. Thus, they are a reasonable parametric approximation of the posterior.

{\bf Dominant mode approximation}. 
Our second approach is based on the intuition that for purposes of prediction in matrix completion tasks, it is sufficient to find only one of the possibly infinitely many solutions to the matrix factorization problem. In this approach, we therefore locate the \emph{dominant mode} from each posterior distribution. We then fit a multivariate Gaussian to the samples correspoding to this mode only, and propagate it as a prior to the following stage. 
 
{\bf Moment matching approximation}.  
Our final approach presents an intermediate between the previous two approaches. Here, we fit a unimodal multivariate Gaussian to the entire set of posterior samples for each parameter using \emph{moment matching}.
Beyond its simplicity, propagating Gaussian approximations for priors has the appeal that the inferences in different stages (i.e. steps \ref{step:infer1}, \ref{step:infer2a}, \ref{step:infer2b}, \ref{step:infer3} in Algorithm \ref{alg:BMFPP}) can be processed as a standard BMF. It also has the usual interpretation that the log-posterior corresponds to a sum-of-squared-errors objective function with quadratic regularization terms \citep{Salakhutdinov+Mnih:2008a}.
Finally, the moment matching approximation brings our scheme in close relation to recent work on expectation propagation for distributed data \citep{Xu+others:2014,Vehtari+others:2018}, but with only limited-communication and a single pass over the data as in assumed density filtering. 

\subsection{Approximating the product density equation for aggregation}\label{sec:aggregation}

Each subset posterior inference results in a joint distribution for subsets of the parameters $
\X$ and $\W$, approximated by a set of samples. Direct aggregation of these joint distributions using the product density equation (\ref{eq:PP_pde}) is a computationally challenging task. For computational efficiency, and to enable the use of the approximations introduced above, we simplify the task by decoupling the parameters and performing the aggregation by posterior marginals. With parametric approximations for each subset posterior marginal available (steps \ref{step:approx1}, \ref{step:approx2a}, \ref{step:approx2b}, \ref{step:approx3} in {Algorithm \ref{alg:BMFPP}}), we aggregate them  
by multiplying them together and dividing away all multiply counted propagated posteriors. 


We assume that the marginal distributions over the parameter matrices can be factorized along rows into a product of $K$-dimensional distributions, i.e. 
\[
\widehat{p}(\X|\Y) = \prod_{n = 1}^N \widehat{p}(\x_n|\Y),\quad \widehat{p}(\W|\Y) = \prod_{d = 1}^D \widehat{p}(\w_d|\Y).
\] 
The \emph{dominant mode} and \emph{moment matching} approximations both produce unimodal multivariate Gaussian representations for each row of the parameter matrices. 
By the properties of Gaussian distributions, the aggregated posterior for the $n$th row of $\X$ is then obtained as
\begin{align}\label{eq:post_marg_aggregation}
\widehat{p}\left(\x_n|\Y\right) &= \mathcal{N}_K\left(\x_n \mid \hat{\mu}^{*}_{\x_n},\left[\hat{\Lambda}^{*}_{\x_n}\right]^{-1}\right),\\
\hat{\Lambda}^{*}_{\x_n} &= \hat{\Lambda}^{(1)}_{\x_n}+\sum_{j=2}^J\left(\hat{\Lambda}^{(j)}_{\x_n}-\hat{\Lambda}^{(1)}_{\x_n}\right),\nonumber\\ 
 \hat{\mu}^{*}_{\x_n} & = \left[\hat{\Lambda}^{*}_{\x_n}\right]^{-1} 
 \left(\hat{\Lambda}^{(1)}_{\x_n}\hat{\mu}^{(1)}_{\x_n}+\sum_{j=2}^J \left( \hat{\Lambda}^{(j)}_{\x_n}\hat{\mu}^{(j)}_{\x_n}-\hat{\Lambda}^{(1)}_{\x_n}\hat{\mu}^{(1)}_{\x_n}\right)\right),\nonumber
\end{align}
where $\hat{\mu}^{(j)}_{\x_n}, \hat{\Lambda}^{(j)}_{\x_n},\;j = 1,\ldots,J$, denote the estimated statistics of the posterior for the $j$th submodel. Note that for submodels indexed by $j=2,\ldots,J$, the effect of the first submodel $(j=1)$ has been removed. The aggregation of each $\widehat{p}(\w_d|\Y)$ is done in similar fashion.

For the GMM approach, the posterior marginal for each $\w_{d}$ and $\x_{n}$ is a mixture with density $f(x)=\sum_{c} \hat{\pi}_{c}\cdot \N\left(\x_{n}; \hat{\mu}^{c}_{\x_{n}}, [\hat{\Lambda}^{c}_{\x_{n}}]^{-1}\right)$. 
%
Maintaining this approximation in the aggregation phase would lead to the computationally challenging problem of dividing one mixture by another. 
Emphasizing speed and efficiency, we instead apply Equation~(\ref{eq:post_marg_aggregation}) using pooled mixture components:
%
\begin{align*}
\hat{\mu}^{(j)}_{\x_{n}} &=\sum^{C}_{c=1} \hat{\pi}_{c}\cdot\hat{\mu}^{c}_{\x_{n}} \label{eq:marginal_mixture} \\
[\hat{\Lambda}^{(j)}_{\x_{n}}]^{-1} &= \sum^{C}_{c=1} \left( \hat{\pi}_{c} [\hat{\Lambda}^{c}_{\x_{n}}]^{-1} + \hat{\pi}_{c}\cdot(\hat{\mu}^{c}_{\x_{n}}-\hat{\mu}^{(j)}_{\x_{n}})(\hat{\mu}^{c}_{\x_{n}}-\hat{\mu}^{(j)}_{\x_{n}})^\top \right). \nonumber
\end{align*}

To improve the numerical stability of using Equation~(\ref{eq:post_marg_aggregation}), we additionally apply an eigenvalue correction to correct for occasionally occurring non-positive definite matrices in the aggregation, which is summarized in Algorithm~\ref{alg:corrected_aggregation}.

\begin{algorithm}[!htb]
\caption{Posterior marginal aggregation of $p(\x_n|\Y)$ using eigenvalue correction. The function \texttt{EigenvalueCorrection}($\mathbf{M}$) finds the smallest eigenvalue of a non-positive definite symmetric matrix $\mathbf{M}$, and adds its absolute value and a small constant to all diagonal elements. The aggregation of $p(\w_d|\Y)$ is done analogously.}\label{alg:corrected_aggregation}
\For {$j = 2:J$ \textbf{in parallel}}{
	\uIf{$\hat{\Lambda}^{(j\setminus 1)}_{\x_n} = \hat{\Lambda}^{(j)}_{\x_n} - \hat{\Lambda}^{(1)}_{\x_n}$ is positive definite}
	{	 $\hat{\Lambda}^{*(j)}_{\x_n}= \hat{\Lambda}^{(j)}_{\x_n}$ }
	\Else {
		$\hat{\Lambda}^{*(j\setminus 1)}_{\x_n} = \text{\texttt{EigenvalueCorrection}}\left(\hat{\Lambda}^{(j\setminus 1)}_{\x_n}\right)$\;	
		$\hat{\Lambda}^{*(j)}_{\x_n} = \hat{\Lambda}^{*(j\setminus 1)}_{\x_n} + \hat{\Lambda}^{(1)}_{\x_n}$
	}
}
$\hat{\Lambda}^{*}_{\x_n} = (2-J) \hat{\Lambda}^{(1)}_{\x_n} + \sum^{J}_{j=2} \hat{\Lambda}^{*(j)}_{\x_n}$\;
$\hat{\mu}^{*}_{\x_n} = \left[ \hat{\Lambda}^{*}_{\x_n} \right]^{-1} \left( (2-J)\hat{\Lambda}^{(1)}_{\x_n} \hat{\mu}^{(1)}_{\x_n} + \sum^{J}_{j=2} \hat{\Lambda}^{*(j)}_{\x_n} \hat{\mu}^{(j)}_{\x_n}\right)$\;
\end{algorithm}

\subsection{Scalability}\label{sec:scalability}

This section provides a brief discussion about the scalability of the above inference scheme in terms of computation time and communication cost. With $U$ workers available, both rows and columns can be partitioned into $\sqrt{U} + 1$ parts, assuming for simplicity an equal number of partitions in both directions (note, however, that this is not a requirement for our method). This results in a total of $U + 2\sqrt{U} + 1$ subsets. The computational cost of a typical BMF inference algorithm per iteration is proportional to $(N+D)K^3 + MK^2$, where $N$ and $D$ are the respective dimensions of the observation matrix, $M$ is the number of observed values, and $K$ is the number of latent dimensions. Thus, for each submodel, the theoretical computation time is proportional to 
\[
t_0:=\left[(N+D)K^3/(\sqrt{U} + 1) + MK^2/(U + 2\sqrt{U} + 1)\right]T,
\] 
assuming an equal number of observations in each subset and $T$ iterations. 
Thus, the initial stage can be completed with one worker in time $t_0$, inference stage II can be processed with $2\sqrt{U}$ workers in time $t_0$, and inference stage III can be completed with $U$ workers in time $t_0$. Finally, the aggregation step mainly involves calculating the product of multivariate Gaussian distributions, which can be done with $2(\sqrt{U} + 1)$ parallel workers in time proportional to
\[
t_{a} := \frac{\max(N, D)}{\sqrt{U} + 1}(K+K^{2}).
\]
Therefore, the total computation time of the algorithm with $U$ worker nodes is proportional to the sum of the computation times of each inference stage plus the computation time of the aggregation, $t = 3 t_0 + t_{a}$.

In terms of communication cost, the proposed inference scheme requires first communicating inputs to workers and then collecting the outputs for aggregation. The inputs consist of two parts: data and prior distributions. As workers use non-overlapping parts of the data, the total amount of communication needed to load the data is proportional to the number of observations $M$. Each worker receives parameters for $(N + D)/(\sqrt{U} + 1)$ distributions, each with $L$ parameters; for the dominant mode and moment matching approximations $L$ is proportional to $K+K^2$ and for the Gaussian mixture model approximation it is proportional to $C(K+K^2)$, where $C$ is the number of components and $K^{2}$ is due to using a full covariance matrix for the posterior of parameters. As there are $U$ workers, the total amount of communication needed for input distributions is proportional to $\sqrt{U}(N + D) L$. The output distributions are of the same size as the input distributions. Thus, the communication cost at the aggregation stage is the same as the communication cost of input distributions.

\section{Experiments}\label{sec:experiments}

In this section, we evaluate the empirical performance of our limited-communication inference scheme for BMF, {\bf posterior propagation with parametric approximations}, by comparing it with both embarrassingly parallel MCMC methods in Section \ref{sec:RMSE_vs_wallClockTime_ep_comparison}, and available state-of-the-art parallel and distributed implementations of BMF and non-negative matrix factorization (NMF) in Section \ref{sec:RMSE_vs_wallClockTime_hpc_implement_comparison}. In Section \ref{sec:subset_posterior_correlation}, we further analyse the advantage of our method over embarrassingly parallel MCMC in terms of encouraging a common representation for model parameters to facilitate subset posterior aggregation. Details about the implementation, test setup and data sets are provided in Sections \ref{sec:implementation}--\ref{sec:datasets}.

\subsection{Implementation}\label{sec:implementation}
For posterior inference in each subset, we use our R implementation\footnote{An implementation of BMF with PP, with highly optimized C libraries, is available in SMURFF software on github (bmfpp branch): \url{https://github.com/ExaScience/smurff}} of the BPMF Gibbs sampler presented by \citet{Salakhutdinov+Mnih:2008b}\footnote{We have corrected an apparent error in the original formulation of the update formula for $[\W_{0}^{*}]^{-1}$, where $\overline{S}$ should be calculated with $\overline{S}=\frac{1}{N} \sum^{N}_{i=1} (\x_{i}-\overline{\X}) (\x_{i}-\overline{\X})^{T}$ \citep{Murphy2007:ConjugateGaussianPrior} rather than $\overline{S}=\frac{1}{N} \sum^{N}_{i=1} \x_{i} \x_{i}^{T}$. Alternatively, one could use the original formula for $\overline{S}$ and calculate $[\W_{0}^{*}]^{-1}$ with $[\W_{0}^{*}]^{-1}= \W_{0}^{-1} + N \overline{S} + \frac{\beta_{0}N}{\beta_{0}+N} (\mu_{0}-\overline{\X}) (\mu_{0}-\overline{\X})^{T} - N \overline{\X} (\overline{\X})^{T}$ or $[\W_{0}^{*}]^{-1}= \W_{0}^{-1} + N \overline{S} + \beta_{0} \mu_{0} (\mu_{0})^{T} - \beta_{0}^{*} \mu_{0}^{*} (\mu_{0}^{*})^{T}$ \citep{Teh2007:ConjugateGaussianPrior}.}. 
BPMF considers the noise precision $\tau$ as a constant and places a normal-Wishart prior on the hyperparameters $\mu_{\W}$ and $\Lambda_{\W}$, as well as on hyperparameters $\mu_{\X}$ and $\Lambda_{\X}$. In the first stage of PP, we sample all parameters of the BPMF model. However, in the second and third stages, we sample hyperparameters $\mu_{\W}$ and $\Lambda_{\W}$ only when the posterior of $\W$ is not propagated. Similarly, hyperparameters $\mu_{\X}$ and $\Lambda_{\X}$ are sampled only when the posterior of $\X$ is not propagated.

We have introduced three alternative approaches (in Section \ref{sec:parametric_approx}) to estimate subset posteriors from samples:
\begin{enumerate}
	\item The GMM approximation fits a mixture of multivariate Gaussians to a clustering of the posterior samples (\textbf{PP GMM}).
	\item The dominant mode approximation fits a multivariate Gaussian to the samples of the dominant mode (\textbf{PP DM}).
	\item The moment matching approximation fits a unimodal multivariate Gaussian to the entire set of posterior samples (\textbf{PP MM}).
\end{enumerate}

%

For a computationally fast way of implementing the first two algorithms, we first use the nonparametric $\lambda$-means clustering algorithm \citep{Comiter+others:2016} to cluster the posterior samples, then (i) for \textbf{PP DM} we choose the cluster with the maximum number of posterior samples to estimate the posterior, (ii) for \textbf{PP GMM} we estimate the posteriors for top-N modes/clusters. In our experiments, we use the top-3 modes.
When using the estimated GMM as a prior for BMF, we perform Gibbs sampling in a similar way as for mixture models; denoting by $\hat{\mu}^{c}_{\x_{n}}$ and $\hat{\Lambda}^{c}_{\x_{n}}$ the estimated mean and precision, respectively, of mode $c$ in the posterior of parameter $\x_n$: 
\begin{enumerate}
	\item Compute the probability $p\left(\x_{n}|\hat{\mu}^{c}_{\x_{n}}, [\hat{\Lambda}^{c}_{\x_{n}}]^{-1}\right)$ of generating the parameter for each mixture component, i.e. the likelihood of $\x_{n}$.
	\item Calculate the responsibility $\gamma\left(\x^{c}_{n}\right)=\hat{\pi}_{c}\cdot p\left(\x_{n}|\hat{\mu}^{c}_{\x_{n}}, [\hat{\Lambda}^{c}_{\x_{n}}]^{-1}\right)$ of each component to explain $\x_{n}$.
	\item Choose the component with the maximum $\gamma\left(\x^{c}_{n}\right)$ as the propagated prior for $\x_{n}$, and update the parameter with its statistics.  
\end{enumerate}
The above procedure is done analogously for $\w_{d}$.

\subsection{Test setup}

We evaluate the distributed inference methods using simulated data and three real-world data sets: {\bf MovieLens-1M}, {\bf MovieLens-20M} and {\bf ChEMBL}. In addition to inference on full data for medium-sized data sets, we predict missing values using column means; this benchmark serves as a baseline and sanity check.

We evaluate performance by predictive accuracy. To this end, we randomly partition the data into training and test sets and use root mean squared error (RMSE) 
on the test set as performance measure. For prediction in the experiments, we use Equation~(\ref{eq:MF}) with posterior means as values for $\X$ and $\W$. 
Furthermore, we use wall-clock time\footnote{Wall-clock time measures the real time between the start and the end of a program. For parallel processes, we use the wall-clock time of the slowest process.} to measure the speed-up achieved by parallelization. 
The reported wall-clock time for our method is calculated by summing the maximum wall-clock times of submodels for each inference stage plus the wall-clock time of the aggregation step\footnote{In our current implementation, we run each stage on a cluster as an array job consisting of multiple independent processes. 
While in this implementation communication between stages is done using read and write operations on disk, its effect on total computation time is negligible as this needs to be done at most three times during the entire progam, and the amount of data communicated is relatively small.
}.


In all experiments, we ran Gibbs sampling with 1200 iterations. We discarded the first 800 samples as burn-in and saved every second of the remaining samples yielding in total 200 posterior samples. The results were averaged over 5 runs. 
For parallelization, we experiment with different partitioning schemes; a partitioning scheme $r\times c$ means that rows are partitioned into $r$ and columns into $c$ subsets. The partitioning scheme $1 \times 1$ refers to the full data. 
Note that the maximum number of parallel workers that can be used by our algorithm is equal to the number of subsets in the third stage, i.e. $(r-1)\times(c-1)$.
We also tested two ordering schemes. In the first scheme, rows and columns are permuted randomly ({\bf row/column order: random}). In the second scheme, the rows and columns are reordered into a descending order according to the proportion of observations in them ({\bf row/column order: decreasing}). Thus, the most dense rows and columns are processed in the first two stages, by which the subsequently propagated posteriors can be made more informative.

The configuration of compute nodes that we used to run the experiments in Subsections \ref{sec:RMSE_vs_wallClockTime_ep_comparison}--\ref{sec:RMSE_vs_wallClockTime_hpc_implement_comparison} is given in Appendix \ref{sc:compute_node_config}.

\subsection{Data sets}\label{sec:datasets}

The {\bf MovieLens-1M} \citep{Harper+Konstan:2015} data set consists of 1,000,209 movie ratings by 6,040 users on 3,706 movies. Approximately 4.5\% of the elements of the movie rating matrix, where each user corresponds to a row and each movie to a column, are observed. The {\bf MovieLens-20M} \citep{Harper+Konstan:2015} data set contains 20 million ratings from 138,493 users on 27,278 movies; that is, about 0.53\% of the elements are observed.
Following \cite{Simm+others:2015}, 
we set $\tau = 1.5$ and $K=10$ for performance analysis.

The {\bf ChEMBL} \citep{Bento+others:2014} data set describes interactions between drugs and proteins using the pIC50 measure. The data set has 15,703 rows corresponding to drugs and 346 columns corresponding to proteins, and contains 59,280 observations which is slightly over 1\% of the elements.
Again, we follow \cite{Simm+others:2015} to set $\tau=5$ and $K=10$. As ChEMBL contains only 346 columns, we only partitioned the rows.

For these real-world data sets, we conduct a 5-fold cross-validation study where 20\% of observations are withheld and used as test set. 

\begin{figure*}[!htb]
\begin{center}
\subfigure[Row/column order: decreasing]{\includegraphics[width=0.925\textwidth]{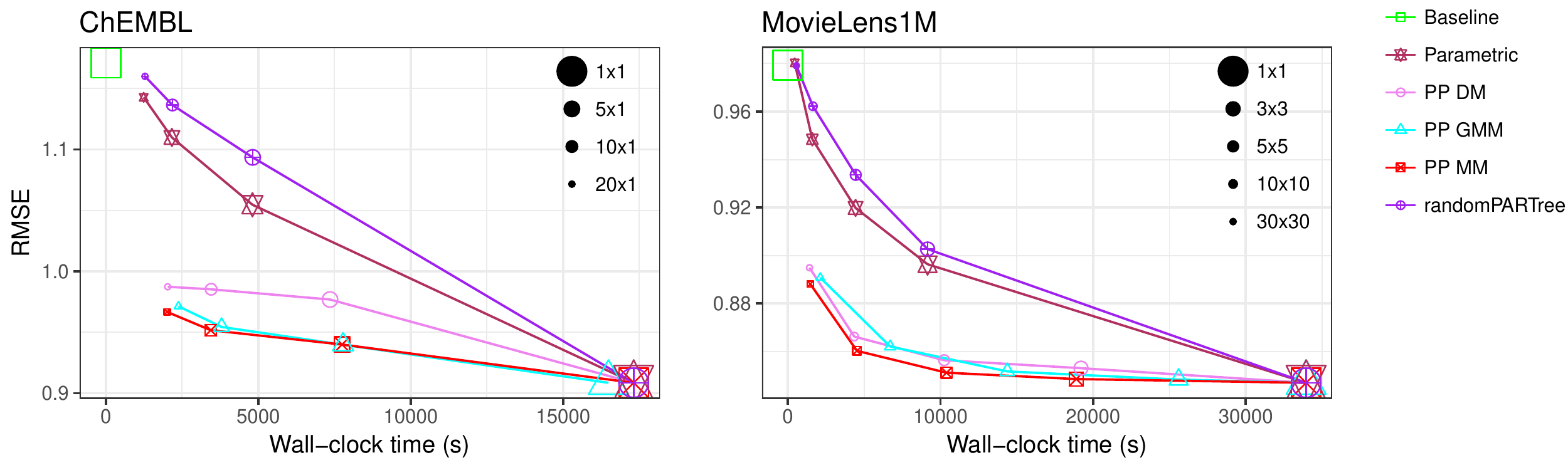}}\\
\subfigure[Row/column order: random]{\includegraphics[width=0.925\textwidth]{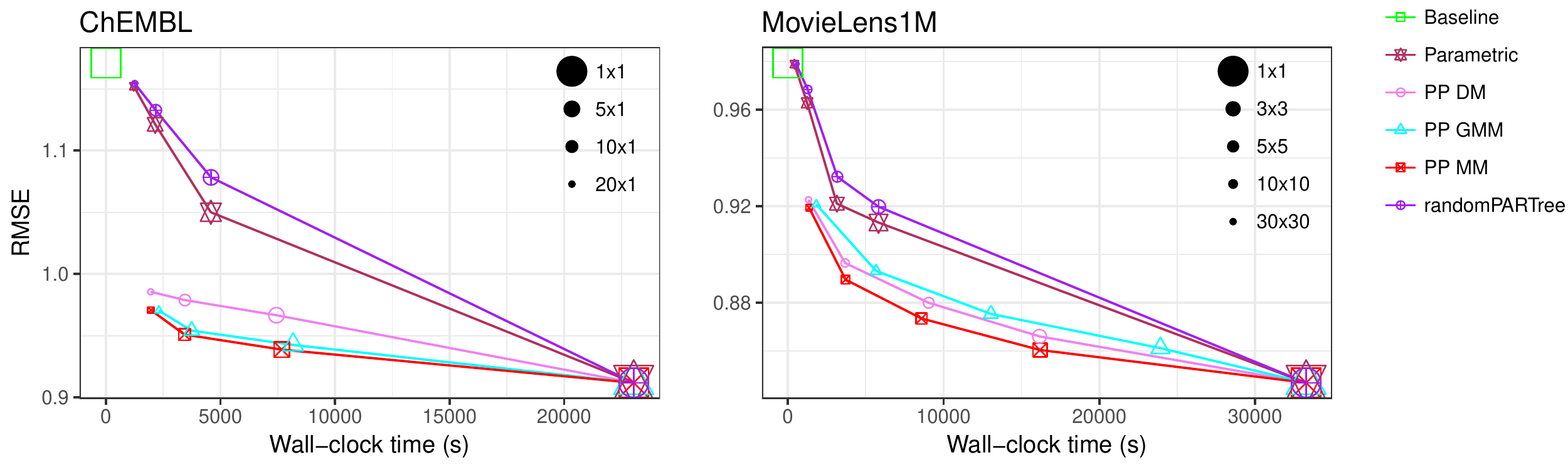}}\\
\subfigure[Row/column order: decreasing]{\includegraphics[width=0.7\textwidth]{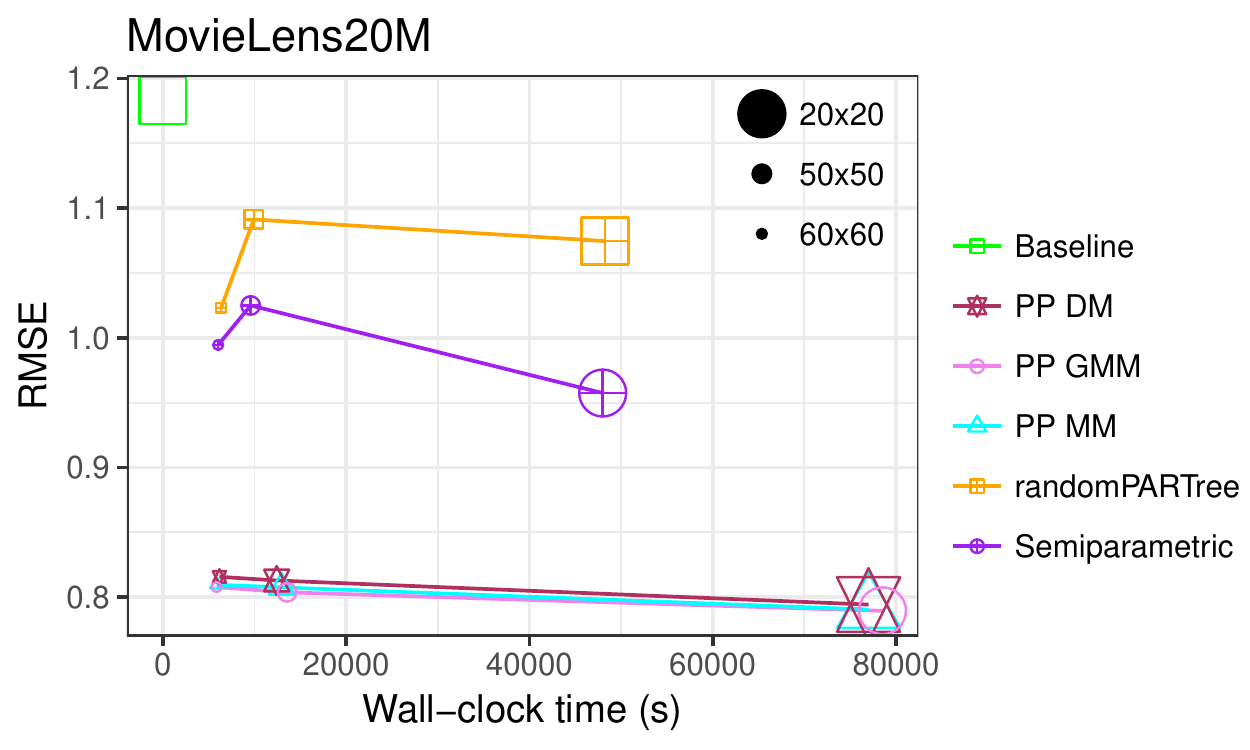}}
\end{center}
\caption{Test RMSE and wall-clock time on ChEMBL (left) and MovieLens-1M (right) data in (a-b), and MovieLens-20M data in (c) with $K=10$. The size of legends/symbols indicates different partition schemes. Lower RMSE is better. {\bf PP MM} works best for all three data sets, followed by {\bf PP GMM} and {\bf PP DM}.}
\label{fig:real_data}
\end{figure*}

To complement the real data sets, we generated {\bf simulated data} sets with 6,040 observations and 3,706 features as follows: We set the number of latent factors to $K=5$. The elements of the matrices $\W$ and $\X$ were generated independently from the standard univariate normal distribution. Finally, we generated the data with $\Y = \X \W^\top + \eps$, where the $\eps$ is a noise matrix whose elements were generated from a standard normal distribution. For learning, we set the parameters $K$ and $\tau$ to the corresponding correct values, i.e., $K=5$ and $\tau=1$. We generated 5 independent simulated data sets.

In many real-world applications, such as collaborative filtering and the ChEMBL benchmark, the data are very sparsely observed. We analyse the predictive performance of the model with respect to different types of missing data. To this end, we randomly select 80\% of the data as missing, use these missing data as test set and the remaining data as training set. To additionally simulate not-missing-at-random data as the second simulated data scenario, we first assigned weights $w_n$ and $w_d$ to each row and column, respectively, such that they form an equally spaced decreasing sequence $0.9,\ldots,0.005$. Then we assigned the element $y_{nd}$ to the test data with probability $w_n w_d$; this results in a matrix with about 80\% of elements missing. This is referred to as the structured missingness scenario.

\subsection{Comparison with embarrassingly parallel methods}\label{sec:RMSE_vs_wallClockTime_ep_comparison}

In this subsection, we compare the predictive performance and computation times of the proposed inference scheme to those of the full model, as well as the following algorithms for embarrassingly parallel MCMC\footnote{We found that running BMF with Gibbs sampler for each subset using down-weighted priors $p(\X)^{1/I}$ and $p(\W)^{1/J}$ could lead to numerical instabilities (i.e. resulting in non-positive definite precision matrices). Therefore, we chose to run BMF for each subset with the standard normal priors $p(\X)$ and $p(\W)$, and then divided away the multiply-counted priors when aggregating the subset posteriors.}:

\begin{enumerate}
	\item {\bf Parametric density product} (parametric)  is a multiplication of Laplacian approximations to subset posteriors. The aggregated samples are drawn from the resulting multivariate Gaussian \citep{Neiswanger+others:2014}.
	\item {\bf Semiparametric density product} (semiparametric) draws aggregated samples from multiplicated semi-parametric estimates of subset posteriors \citep{Neiswanger+others:2014}.
	\item {\bf Random partition tree} (randomPARTree) works by first performing space partitioning over subset posteriors, followed by a density aggregation step which simply multiplies densities across subsets for each block and then normalizes \citep{Wang+others:2015}. Aggregated samples are drawn from the aggregated posterior.
\end{enumerate}
All of the above algorithms are implemented in the Matlab PART library\footnote{\url{https://github.com/wwrechard/random-tree-parallel-MCMC}}. We ran the randomPARTree algorithm with different values for its hyperparameters (i.e. min cut length=0.001 or 0.01, min fraction block=0.01 or 0.1, cut type=``kd" or ``ml", local gaussian smoothing=true or false) for pilot analysis, and found that there are no significant differences in the predictive performance for different values for the hyperparameters. Thus, for this algorithm, we use KD-tree for space partition and 1000 resamples for final approximation, and use the default values for the other hyperparameters provided in the library.

\subsubsection{Results}
\begin{figure*}[!htb]
\begin{center}
\subfigure[Row/column order: decreasing]{\includegraphics[width=0.925\textwidth]{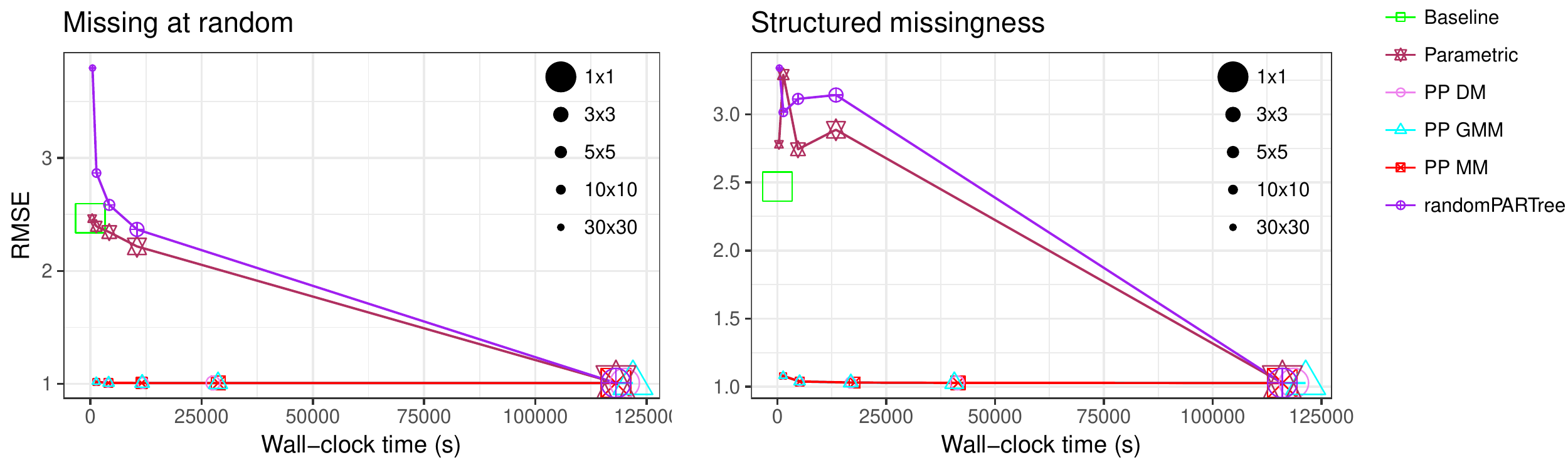}}\\
\subfigure[Row/column order: random]{\includegraphics[width=0.925\textwidth]{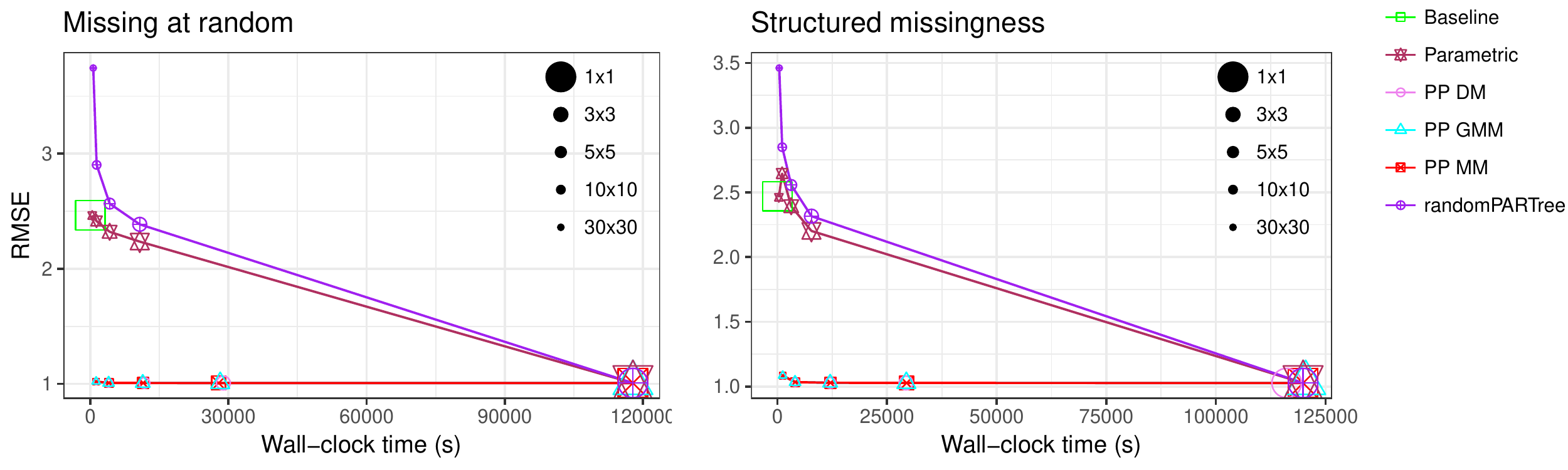}}
\end{center}
\caption{Test RMSE and wall-clock time on simulated data with 80$\%$ missing values. Lower RMSE is better. Left: values missing at random. Right: missing data generated with the structured missingness scenario. Nearly horizontal curves imply that posterior propagation speeds up computation with almost no loss on accuracy.}
\label{fig:simulated_data}
\end{figure*}

The results for ChEMBL, MovieLens-1M, MovieLens-20M, and simulated data are shown in Figures~\ref{fig:real_data} and \ref{fig:simulated_data}. To improve the readability of the plots, we only plot RMSE for the two posterior aggregation methods that give the best performance for embarrassingly parallel MCMC on each data set. 
In the following, we summarize the major conclusions of this evaluation.

\begin{figure}[!htb]
\begin{center}
\subfigure{\includegraphics[page=1,width=0.925\textwidth]{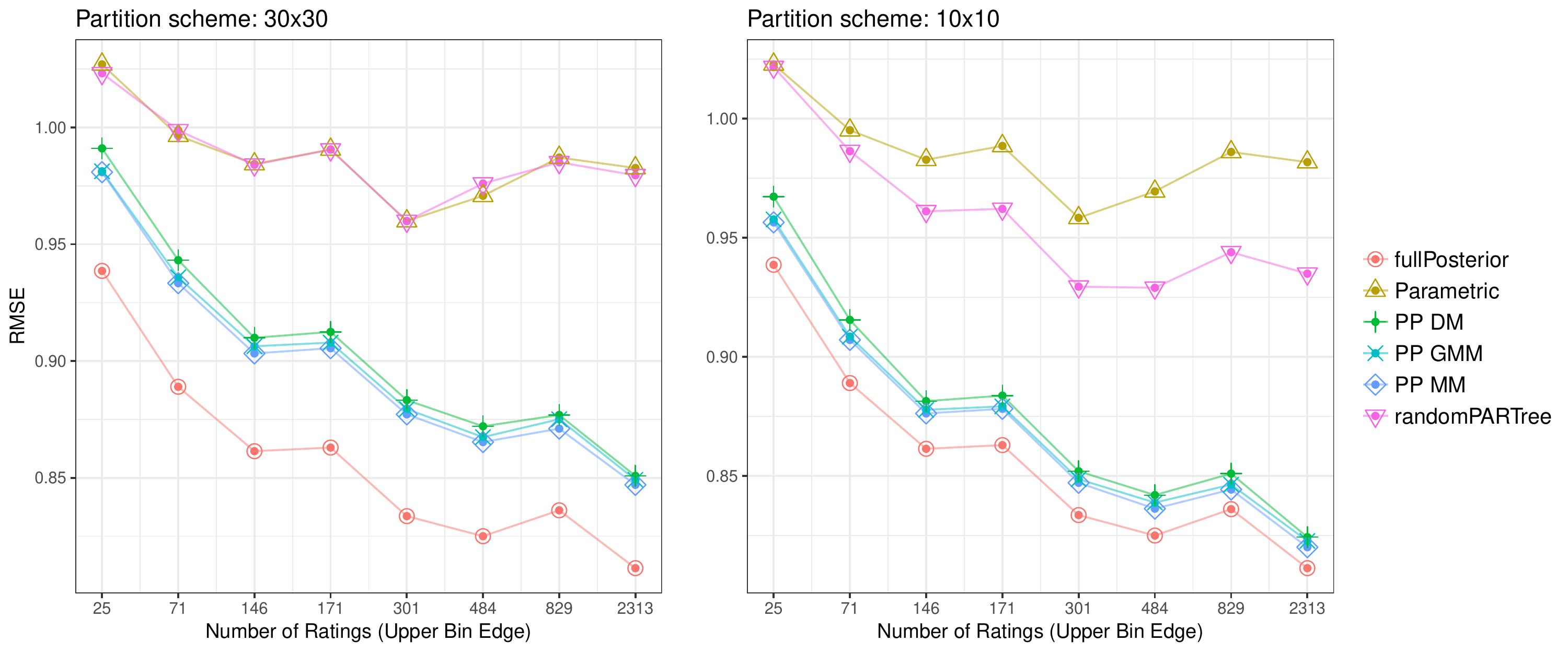}}\\
\subfigure{\includegraphics[page=2,width=0.925\textwidth]{figs/MovieLen_1m_BMF_RMSE_vs_ratingBins.pdf}}
\end{center}
\caption{Test RMSE with respect to user frequency (i.e. the number of ratings the users have in the training set) on MovieLens-1M (row/column order: decreasing) data with K = 10 for different partition schemes. Here the fullPosterior refers to the posterior on the full data (i.e. partition scheme $1 \times 1$). Compared with the embarrassingly parallel methods, our proposed methods can produce predictions with lower variance for all user groups and produce predictions comparable with that of the full posterior if the size of the subset is large enough (e.g. subset with 1,208 rows by 742 columns for partition scheme $5 \times 5$).}
\label{fig:rmse_vs_ratingBins}
\end{figure}

As a general conclusion, we found that posterior propagation can give almost an order of magnitude faster computation times with a negligible effect on predictive accuracy, compared to MCMC inference on the full data matrix; this can be seen on simulated data and MovieLens (Figure~\ref{fig:simulated_data} and right-hand side of Figure~\ref{fig:real_data} (a)). The almost horizontal RMSE curves for our methods on MovieLens-20M and simulated data indicate that posterior propagation speeds up computation with almost no loss on accuracy. Note that without approximations, PP would give the same results as the full model. The difference between them therefore \emph{quantifies the effect of the approximations} made in our approach. 

Of the embarrassingly parallel MCMC methods, the parametric aggregation method gives the best predictive accuracy on ChEMBL and MovieLens-1M data. Posterior propagation provides better predictive accuracy (lower RMSE values) than any of the embarrassingly parallel MCMC methods on all of the data sets considered. 
We also found out that reordering rows and columns, in a decreasing order with respect to the number of observations, usually improves the accuracy of posterior propagation compared to using a random order of rows and columns; this can be seen on the sparsely observed MovieLens-1M data (right-hand side of Figure~\ref{fig:real_data} (a,b)).
In Appendix \ref{sec:run_time_speedup}, we analyse the results in Figures~\ref{fig:real_data} and \ref{fig:simulated_data} from another perspective to show the wall-clock time speed-up as a function of the number of parallel workers.


We further explored empirically whether posterior propagation can produce good prediction for users and items with only a few observations. This is useful for cold-start problems, i.e., recommendation for new users with very few observed ratings. For this analysis, we visualize test RMSE versus the number of ratings per user in the training set in Figure \ref{fig:rmse_vs_ratingBins}. Again, we observed that compared with the alternative embarrassingly parallel MCMC methods, our methods show superior performance for all user groups and improve prediction for users with very few observed ratings.

\subsection{Comparison with other parallel and distributed implementations}\label{sec:RMSE_vs_wallClockTime_hpc_implement_comparison}

In this subsection, we show that our method achieves competitive results compared
to alternative implementations of parallel and distributed BMF, while keeping the communication requirement bounded. To this end, we compare our method on large-scale data (MovieLens-20M) with the following implementations:
\begin{enumerate}
    \item \textbf{Distributed parallel BPMF}\footnote{\url{https://github.com/ExaScience/bpmf}} (D-BPMF): a state-of-the-art C++ implementation of distributed BPMF with Gibbs sampler \citep{VanderAa+others:2017}. It supports hybrid communication for distributed learning, which utilizes TBB for shared memory level parallelism and Global Address Space Programming Interface (GASPI) for cross-node communication.
    \item \textbf{NMF with parallel SGLD}\footnote{\url{https://perso.telecom-paristech.fr/simsekli/nmf_sgmcmc/}} (NMF + PSGLD): OpenMP implementation of non-negative matrix factorization 
    with parallel SGLD \citep{Simsekli+others:2017}. This is an open source software that is similar to BPMF with distributed SGLD\footnote{The software for large scale BPMF with distributed SGLD is not publicly available.} \citep{Ahn+others:2015}. This software requires careful tuning of hyperparameters in order to avoid numerical instabilities/overflow issues and get reasonable predictions. We set $\epsilon=0.0001$, $\beta=2$ (using a Gaussian likelihood), $\lambda=0.01$, $initStd=0.5$ for the experiment based on a pilot study.
\end{enumerate}

For our method, we used a $20 \times 20$ partition scheme with the same setup as for the experiments in Figure \ref{fig:real_data}(c): i.e. $K=10$, $T=1200$ iterations for Gibbs sampling and a burn-in of 800 samples. Note that our method was implemented in the R language without any optimization, while the other two methods were implemented with highly optimized C libraries. For the sake of obtaining comparable results, the RMSE was obtained using our R implementation (same as in Figure \ref{fig:real_data}(c)), while the wall-clock time is an estimate computed by using 
D-BPMF within each individual data subset in the three stages of our posterior propagation scheme. For each subset, we used a single node with 24 cores, resulting in a total of $(20-1)^2=361$ parallel processes (one per node) for the third stage. 

Since only the OpenMP implementation of NMF + PSGLD was available to us, it was run within a single compute node with no cross-node communication.

\subsubsection{Results}

The results of the comparison are shown in Table~\ref{Tab:comparison_mpi_methods}.
The approximations made in our approach (BMF + PP) have only a slight negative effect on the RMSE compared to D-BPMF, which makes no distributional approximations and should therefore be close to the accuracy of the full model (not computed here because of its large size). 
On the other hand, in terms of computation time, our method is able to leverage 
the combination of a high level of parallelism and a very low communication frequency
compared to D-BPMF, which requires frequent communication.
Varying the number of nodes / cores for the latter yields results which are consistent with the empirical finding of \citet{VanderAa+others:2017}: parallel efficiency initially improves with increased parallelism, but begins to level off as the number of nodes increases beyond the boundary of fast network connections in the HPC cluster. 
A further point to note is that the communication cost of 
D-BPMF increases linearly with respect to the number of MCMC samples while ours stays constant, thus, the longer chains we run, the more advantage we get.  
Finally, NMF + PSGLD performs worse than the other alternative methods in terms of predictive accuracy and wall-clock computation time. This is partially due to the difficulty of tuning the hyperparameters for each specific data set for the DSGLD and PSGLD methods.  

\begin{table}[!htb]
\caption{Comparison of performance for different implementations of distributed BMF / NMF on MovieLens-20M data. The communication costs are computed using the formulas in Table~\ref{Tab:comparison_communication_cost} and consider cross-node (distributed memory) communication only. Note that for D-BPMF, communication takes place in every iteration, whereas for the proposed method communication is only required between stages of inference. 
Also note that in practice the factor $(K+K^2)$ in the cost of the proposed method reduces to $K(K+3)/2$ due to the precision matrix being symmetric.
}
\label{Tab:comparison_mpi_methods}
\renewcommand{\arraystretch}{1.25}
\begin{adjustwidth}{-.5in}{-.5in}
\begin{center}
 \begin{tabular}{l|c|c|c|c}
  \hline
\multirow{2}{*}{Model} & \multirow{2}{*}{\#Nodes/\#cores} & \multirow{2}{*}{RMSE} & \multirow{2}{*}{Wall-clock time (s)} & Communication cost \\
& & & & (floating-point unit)\\ \hline
Proposed method & \multirow{2}{*}{361/8664} & \multirow{2}{*}{0.789} & \multirow{2}{*}{\bf 70} & \multirow{2}{*}{$3.15 \cdot 10^{6} \cdot (K+K^{2})$} \\ 
(BMF + PP) &  &  &  &  \\ \hline

\multirow{5}{*}{D-BPMF} & 1/24 & 0.779 & 204 & - \\
 & $2/48$ & 0.781 & 158 & $1.65 \cdot 10^{5} \cdot K \cdot T$ \\ 
 & $8/192$ & 0.780 & 89 & $1.16 \cdot 10^{6} \cdot K \cdot T$ \\ 
\citep{VanderAa+others:2017}  & 64/1536 & {\bf 0.772} & 124 & $1.04 \cdot 10^{7} \cdot K \cdot T$ \\ 
  & 85/2040 & 0.780 & 120 & $1.39 \cdot 10^{7} \cdot K \cdot T$ \\ 
\hline

NMF + PSGLD  & \multirow{2}{*}{1/24} & \multirow{2}{*}{0.903} & \multirow{2}{*}{9837} & \multirow{2}{*}{-} \\ 
\citep{Simsekli+others:2017} &  &  &  &  \\ 

\hline
\end{tabular}
\end{center}
\end{adjustwidth}
\end{table}

\subsection{Correlations of the subset posteriors}\label{sec:subset_posterior_correlation}

We observed in Section \ref{sec:RMSE_vs_wallClockTime_ep_comparison} that compared with existing embarrassingly parallel methods, the proposed method can provide a better trade-off between predictive accuracy and wall-clock time on several benchmark data sets. In this section, we investigate empirically to what extent our method is able to encourage joint identifiability via dependencies between the inferences in different subsets.

\begin{figure*}[!htb]
\begin{center}

\subfigure{\includegraphics[page=1,width=1.0\textwidth]{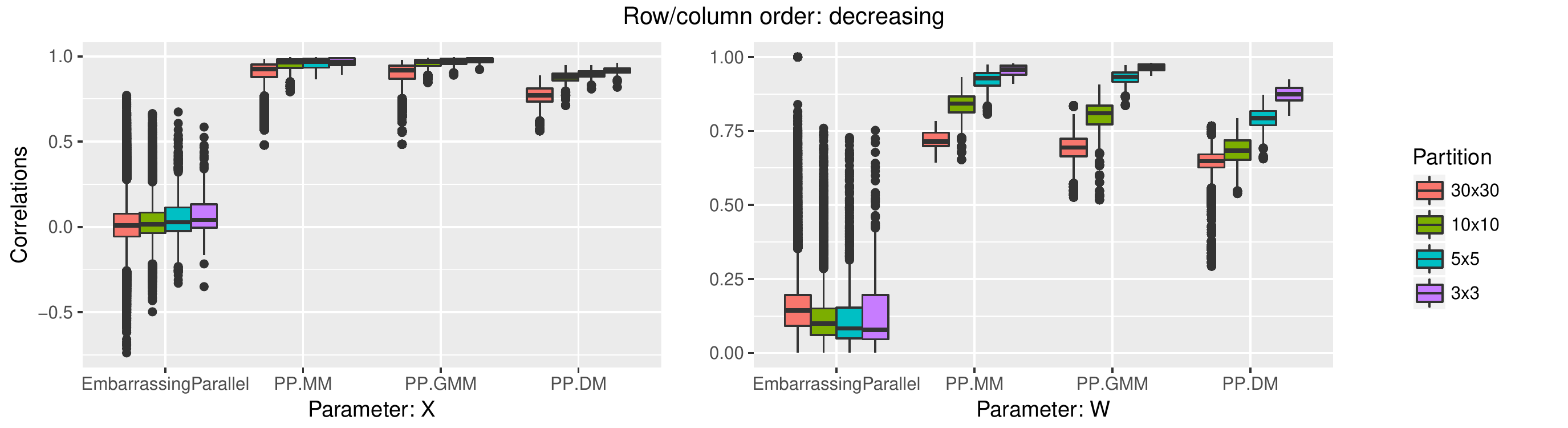}}\\
\subfigure{\includegraphics[page=2,width=1.0\textwidth]{figs/MovieLen_1m_submodel_posteriorMean_correlationStats.pdf}}

\end{center}
\caption{The correlations of posterior estimates of different subsets for different variants of the proposed method and embarrassingly parallel MCMC on MovieLens-1M data with K = 10, for different partition schemes. Compared to embarrassingly parallel MCMC, the proposed method can produce posterior estimates which are highly correlated between different subsets, suggesting that it can enforce a common representation for model parameters, making the aggregation of submodels feasible.} 

\label{fig:correlation_subposteriors_movielens1m}
\end{figure*}

For this purpose, we compute for our method the correlations between the posterior expected means of parameters in subsets sharing rows or columns, and compare them to the corresponding correlations produced by embarrassingly parallel MCMC, see Figure \ref{fig:correlation_subposteriors_movielens1m}. 
For example, for the partition scheme in Figure \ref{fig:PP}, we would calculate the correlations of posterior means for subsets as follows: cor($\hat{\X}^{(1,1)}$, $\hat{\X}^{(1,j)}$), cor($\hat{\X}^{(i,1)}$, $\hat{\X}^{(i,j)}$), cor($\hat{\W}^{(1,1)}$, $\hat{\W}^{(i,1)}$), cor($\hat{\W}^{(1,j)}$, $\hat{\W}^{(i,j)}$), for $i=2, \cdots, I$, $j=2, \cdots, J$. 
For embarrassingly parallel MCMC, to avoid low correlations due to mis-aligned permutations of the latent dimensions in different submodels, highly correlated latent dimensions were aligned prior to calculating the correlations between the posterior means. 

An obvious trend from Figure \ref{fig:correlation_subposteriors_movielens1m} is that the correlation scores of posterior estimates generated by our method are much higher than those of embarrassingly parallel MCMC. The observation suggests that by propagating the posteriors obtained from the earlier stage to the next stage as priors, our method can produce highly dependent subset posteriors. 
On the other hand, since existing embarrassingly parallel MCMC methods do not introduce any dependencies between the inferences for different subsets, they are unable to enforce a common permutation and scaling  for parameters, making the aggregation step challenging for unidentifiable models.

\section{Discussion}

We have introduced a hierarchical embarrassingly parallel strategy for Bayesian matrix factorization, which enables a trade-off between accuracy and computation time, and uses very limited communication. The empirical evaluation on both real and simulated data shows that (i) our distributed approach is able to achieve a speed-up of almost an order-of-magnitude, with a negligible effect on predictive accuracy, compared to MCMC inference on the full data matrix; (ii) our method also significantly outperforms state-of-the-art embarrassingly parallel MCMC methods in accuracy, and (iii) performs comparably to other available distributed and parallel implementations of BMF. We further show that, unlike existing embarrassingly parallel approaches, our method produces posterior estimates, which are highly correlated across different subsets and thus enable a meaningful aggregation of subset inferences. 

We have experimented with both inclusive approximations (GMM and MM; attempting to include all sampled modes, which is still restricted to a small finite number) as well as exclusive approximations (DM; attempting to exclude all but one mode, a property shared with variational inference). In our current setting, the inclusive approximations gave more consistent performance, striking a balance between restricting the set of solutions to encourage identifiability and letting the sampler explore good solutions. 
%
%
While the proposed approximations work well, 
more accurate representations for subset posteriors could be considered instead, in particular for aggregation \citep[e.g.][]{Wang+others:2015}. 
This would be relevant especially if we were interested in an accurate, global representation of the joint distribution of $\X$ and $\W$, instead of predicting unobserved elements of the data matrix $\Y$, which was the case in our current work. While more sophisticated representations may improve accuracy, they come at the expense of increased computational burden. 
An additional motivation to consider alternative subset posterior representations is to be able to establish theoretical guarantees for the global model, which despite their good empirical performance, are not straightforward to give for the current approximations.

Several works on distributed learning for BMF (e.g. BMF with DSGLD \citep{Ahn+others:2015}, NMF with parallel SGLD \citep{Simsekli+others:2017}) assume (implicitly) that the models are trained with blocks in an orthogonal group (or squared partition), in order to avoid conflicting access to parameters among parallel workers. In our work, we do not make any assumptions about the partition scheme, and our method can therefore work flexibly with diversified partition schemes, which depend on the size of the data in both dimensions. For instance, it can work with partitions only along the row direction for tall data, or partitions only along the column direction for fat data, or partitions along both row and column directions for tall and fat data.

The focus of our work has been to develop an efficient and scalable distributed learning scheme for BMF. In doing so, we have assumed implicitly that the data are missing at random, following many other works on BMF. 
While many (if not most) real-world data sets exhibit non-random patterns on missingness, we have handled such patterns using the simple strategy of reordering rows and columns into a descending order according to the proportion of observations in them. Thus, the most dense rows and columns are processed during the first two stages, by which the subsequently propagated posteriors can be made more informative.
However, it is also possible to handle non-random patterns of missing values in a more principled manner. In the context of matrix factorization, \cite{Hernandez-Lobato:2014:PMF:MNAR} modelled the generative process for both the data and the missing data mechanism, and showed empirically that learning these two models jointly can improve performance of the MF model. This strategy would be straightforward to incorporate within our distributed scheme.

Finally, we have run experiments on a scale of only tens of millions of elements, 
but there is no obstacle for running the proposed distributed algorithm on larger matrices. 
%
Indeed, the proposed approach as such is not implementation-dependent, and it could be used together with any available well-optimized (but more communication intensive) implementation to enable further scaling beyond the point at which parallel efficiency would otherwise begin to level off.

\begin{acknowledgements}
This project has received funding from the European Union's Horizon 2020 Research and Innovation programme under Grant Agreement no. 671555 and from the Academy of Finland (Finnish Centre of Excellence in Computational Inference Research COIN and grants 294238, 292334). The authors gratefully acknowledge the computational resources provided by the Aalto Science-IT project and by IT4I under Project ID OPEN-11-20. We also thank Tom Vander Aa for his help in setting up and configuring their distributed parallel BPMF software \citep{VanderAa+others:2017} on the Salomon cluster, and the authors of \citet{Simsekli+others:2017} for sharing their software for NMF + PSGLD. Finally, we wish to thank the four anonymous reviewers for their constructive feedback, which lead to many improvements of the paper.
\end{acknowledgements}

\bibliographystyle{spbasic}      
\bibliography{bibliography}

\begin{thebibliography}{33}
\providecommand{\natexlab}[1]{#1}
\providecommand{\url}[1]{{#1}}
\providecommand{\urlprefix}{URL }
\expandafter\ifx\csname urlstyle\endcsname\relax
  \providecommand{\doi}[1]{DOI~\discretionary{}{}{}#1}\else
  \providecommand{\doi}{DOI~\discretionary{}{}{}\begingroup
  \urlstyle{rm}\Url}\fi
\providecommand{\eprint}[2][]{\url{#2}}

\bibitem[{Adams et~al(2010)Adams, Dahl, and Murray}]{Adams+others:2010}
Adams R, Dahl G, Murray I (2010) Incorporating side information in
  probabilistic matrix factorization with {G}aussian processes. In: Proceedings
  of the 26th Annual Conference on Uncertainty in Artificial Intelligence, AUAI
  Press, pp 1--9

\bibitem[{Ahn et~al(2015)Ahn, Korattikara, Liu, Rajan, and
  Welling}]{Ahn+others:2015}
Ahn S, Korattikara A, Liu N, Rajan S, Welling M (2015) Large-scale distributed
  {B}ayesian matrix factorization using stochastic gradient {MCMC}. In:
  Proceedings of the 21th ACM SIGKDD International Conference on Knowledge
  Discovery and Data Mining, pp 9--18, \doi{10.1145/2783258.2783373}

\bibitem[{Angelino et~al(2016)Angelino, Johnson, and
  Adams}]{Angelino+others:2016}
Angelino E, Johnson MJ, Adams RP (2016) Patterns of scalable {B}ayesian
  inference. Foundations and Trends in Machine Learning 9(2-3):119--247,
  \doi{10.1561/2200000052}

\bibitem[{Bento et~al(2014)Bento, Gaulton, Hersey, Bellis, Chambers, Davies,
  Kr\"uger, Light, Mak, McGlinchey, Nowotka, Papadatos, Santos, and
  Overington}]{Bento+others:2014}
Bento A, Gaulton A, Hersey A, Bellis L, Chambers J, Davies M, Kr\"uger F, Light
  Y, Mak L, McGlinchey S, Nowotka M, Papadatos G, Santos R, Overington J (2014)
  The {ChEMBL} bioactivity database: an update. Nucleic Acids Research
  42(D1):D1083--D1090

\bibitem[{Bhattacharya and Dunson(2011)}]{Bhattacharya+Dunson:2011}
Bhattacharya A, Dunson DB (2011) Sparse {B}ayesian infinite factor models.
  Biometrika 98(2):291--306, \doi{10.1093/biomet/asr013}

\bibitem[{Cobanoglu et~al(2013)Cobanoglu, Liu, Hu, Oltvai, and
  Bahar}]{Cobanoglu+others:2013}
Cobanoglu MC, Liu C, Hu F, Oltvai ZN, Bahar I (2013) Predicting drug-target
  interactions using probabilistic matrix factorization. Journal of Chemical
  Information and Modeling 53(12):3399--3409, \doi{10.1021/ci400219z}

\bibitem[{Comiter et~al(2016)Comiter, Cha, Kung, and
  Teerapittayanon}]{Comiter+others:2016}
Comiter M, Cha M, Kung HT, Teerapittayanon S (2016) Lambda means clustering:
  Automatic parameter search and distributed computing implementation. In: 2016
  23rd International Conference on Pattern Recognition (ICPR), pp 2331--2337,
  \doi{10.1109/ICPR.2016.7899984}

\bibitem[{\c{S}im\c{s}ekli et~al(2015)\c{S}im\c{s}ekli, Koptagel,
  G{\"u}lda\c{s}, Cemgil, {\"O}ztoprak, and Birbil}]{Simsekli+others:2015}
\c{S}im\c{s}ekli U, Koptagel H, G{\"u}lda\c{s} H, Cemgil AT, {\"O}ztoprak F,
  Birbil {\c{S}}I (2015) Parallel stochastic gradient {MCMC} for matrix
  factorisation models. arXiv preprint arXiv:150601418

\bibitem[{\c{S}im\c{s}ekli et~al(2017)\c{S}im\c{s}ekli, Durmus, Badeau,
  Richard, Moulines, and Cemgil}]{Simsekli+others:2017}
\c{S}im\c{s}ekli U, Durmus A, Badeau R, Richard G, Moulines {\'E}, Cemgil AT
  (2017) Parallelized stochastic gradient {M}arkov chain {M}onte {C}arlo
  algorithms for non-negative matrix factorization. In: {IEEE} International
  Conference on Acoustics, Speech and Signal Processing, {ICASSP} 2017, pp
  2242--2246

\bibitem[{Harper and Konstan(2015)}]{Harper+Konstan:2015}
Harper FM, Konstan JA (2015) The {MovieLens} datasets: History and context. ACM
  Transactions on Interactive Intelligent Systems 5(4)

\bibitem[{Hern\'{a}ndez-Lobato et~al(2014)Hern\'{a}ndez-Lobato, Houlsby, and
  Ghahramani}]{Hernandez-Lobato:2014:PMF:MNAR}
Hern\'{a}ndez-Lobato JM, Houlsby N, Ghahramani Z (2014) Probabilistic matrix
  factorization with non-random missing data. In: Proceedings of the 31st
  International Conference on International Conference on Machine Learning -
  Volume 32, pp II--1512--II--1520

\bibitem[{Koren et~al(2009)Koren, Bell, and Volinsky}]{Koren+others:2009}
Koren Y, Bell R, Volinsky C (2009) Matrix factorization techniques for
  recommender systems. Computer 42(8):30--37, \doi{10.1109/MC.2009.263}

\bibitem[{Lian et~al(2015)Lian, Huang, Li, and
  Liu}]{Lian:2015:APS:threadSpeedup}
Lian X, Huang Y, Li Y, Liu J (2015) Asynchronous parallel stochastic gradient
  for nonconvex optimization. In: Proceedings of the 28th International
  Conference on Neural Information Processing Systems - Volume 2, pp 2737--2745

\bibitem[{Lopes and West(2004)}]{Lopes+West:2004}
Lopes HF, West M (2004) Bayesian model assessment in factor analysis.
  Statistica Sinica 14(1):41--67

\bibitem[{Minsker et~al(2014)Minsker, Srivastava, Lin, and
  Dunson}]{Minsker+others:2014}
Minsker S, Srivastava S, Lin L, Dunson D (2014) Robust and scalable {B}ayes via
  a median of subset posterior measures. arXiv preprint arXiv:14032660

\bibitem[{Murphy(2007)}]{Murphy2007:ConjugateGaussianPrior}
Murphy KP (2007) {Conjugate {B}ayesian analysis of the {G}aussian
  distribution}. Tech. rep., Google

\bibitem[{Neiswanger et~al(2014)Neiswanger, Wang, and
  Xing}]{Neiswanger+others:2014}
Neiswanger W, Wang C, Xing E (2014) Asymptotically exact, embarrassingly
  parallel {MCMC}. In: Proceedings of the 30th Conference on Uncertainty in
  Artificial Intelligence, {AUAI} Press, pp 623--632

\bibitem[{Nemeth and Sherlock(2017)}]{Nemeth+Sherlock:2017}
Nemeth C, Sherlock C (2017) Merging {MCMC} subposteriors through
  {G}aussian-process approximations. Bayesian Analysis \doi{10.1214/17-BA1063},
  advance publication

\bibitem[{Park et~al(2013)Park, Kim, and Choi}]{Park+others:2013}
Park S, Kim YD, Choi S (2013) Hierarchical {B}ayesian matrix factorization with
  side information. In: Proceedings of the 23rd International Joint Conference
  on Artificial Intelligence, AAAI Press, pp 1593--1599

\bibitem[{Porteous et~al(2010)Porteous, Asuncion, and
  Welling}]{Porteous+others:2010}
Porteous I, Asuncion A, Welling M (2010) Bayesian matrix factorization with
  side information and {D}irichlet process mixtures. In: Proceedings of the
  24th AAAI Conference on Artificial Intelligence, AAAI Press, pp 563--568

\bibitem[{Salakhutdinov and
  Mnih(2008{\natexlab{a}})}]{Salakhutdinov+Mnih:2008b}
Salakhutdinov R, Mnih A (2008{\natexlab{a}}) Bayesian probabilistic matrix
  factorization using {M}arkov chain {M}onte {C}arlo. In: Proceedings of the
  25th International Conference on Machine Learning, ACM, pp 880--887,
  \doi{10.1145/1390156.1390267}

\bibitem[{Salakhutdinov and
  Mnih(2008{\natexlab{b}})}]{Salakhutdinov+Mnih:2008a}
Salakhutdinov R, Mnih A (2008{\natexlab{b}}) Probabilistic matrix
  factorization. In: Advances in Neural Information Processing Systems 20, MIT
  Press

\bibitem[{Scott et~al(2016)Scott, Blocker, Bonassi, Chipman, George, and
  McCulloch}]{Scott+others:2016}
Scott SL, Blocker AW, Bonassi FV, Chipman HA, George EI, McCulloch RE (2016)
  Bayes and big data: the consensus {M}onte {C}arlo algorithm. International
  Journal of Management Science and Engineering Management 11(2):78--88,
  \doi{10.1080/17509653.2016.1142191}

\bibitem[{Simm et~al(2015)Simm, Arany, Zakeri, Haber, Wegner, Chupakhin,
  Ceulemans, and Moreau}]{Simm+others:2015}
Simm J, Arany A, Zakeri P, Haber T, Wegner JK, Chupakhin V, Ceulemans H, Moreau
  Y (2015) Macau: Scalable {B}ayesian multi-relational factorization with side
  information using {MCMC}. arXiv preprint arXiv:150904610

\bibitem[{Srivastava et~al(2015)Srivastava, Li, and
  Dunson}]{Srivastava+others:2015}
Srivastava S, Li C, Dunson D (2015) Scalable {B}ayes via barycenter in
  {W}asserstein space. arXiv preprint arXiv:150805880

\bibitem[{Teh(2007)}]{Teh2007:ConjugateGaussianPrior}
Teh YW (2007) {Exponential {F}amilies: {G}aussian, {G}aussian-{G}amma,
  {G}aussian-{W}ishart, {M}ultinomial}. Tech. rep., University College London

\bibitem[{{Vander Aa} et~al(2016){Vander Aa}, Chakroun, and
  Haber}]{vander2016distributed}
{Vander Aa} T, Chakroun I, Haber T (2016) Distributed {B}ayesian probabilistic
  matrix factorization. In: 2016 {IEEE} International Conference on Cluster
  Computing, {IEEE} Computer Society, pp 346--349

\bibitem[{{Vander Aa} et~al(2017){Vander Aa}, Chakroun, and
  Haber}]{VanderAa+others:2017}
{Vander Aa} T, Chakroun I, Haber T (2017) Distributed {B}ayesian probabilistic
  matrix factorization. Procedia Computer Science 108:1030 -- 1039

\bibitem[{Vehtari et~al(2018)Vehtari, Gelman, Sivula, Jyl\"{a}nki, Tran, Sahai,
  Blomstedt, Cunningham, Schiminovich, and Robert}]{Vehtari+others:2018}
Vehtari A, Gelman A, Sivula T, Jyl\"{a}nki P, Tran D, Sahai S, Blomstedt P,
  Cunningham JP, Schiminovich D, Robert C (2018) Expectation propagation as a
  way of life: A framework for {B}ayesian inference on partitioned data. arXiv
  preprint arXiv:14124869 \urlprefix\url{https://arxiv.org/abs/1412.4869}

\bibitem[{Wang and Dunson(2013)}]{Wang+Dunson:2013}
Wang X, Dunson D (2013) Parallelizing {MCMC} via {W}eierstrass sampler. arXiv
  preprint arXiv:13124605

\bibitem[{Wang et~al(2014)Wang, Peng, and Dunson}]{Wang+others:2014}
Wang X, Peng P, Dunson DB (2014) Median selection subset aggregation for
  parallel inference. In: Advances in {N}eural {I}nformation {P}rocessing
  {S}ystems 27, Curran Associates, Inc., pp 2195--2203

\bibitem[{Wang et~al(2015)Wang, Guo, Heller, and Dunson}]{Wang+others:2015}
Wang X, Guo F, Heller KA, Dunson DB (2015) Parallelizing {MCMC} with random
  partition trees. In: Advances in Neural Information Processing Systems 28,
  Curran Associates, Inc., pp 451--459

\bibitem[{Xu et~al(2014)Xu, Lakshminarayanan, Teh, Zhu, and
  Zhang}]{Xu+others:2014}
Xu M, Lakshminarayanan B, Teh YW, Zhu J, Zhang B (2014) Distributed {B}ayesian
  posterior sampling via moment sharing. In: Ghahramani Z, Welling M, Cortes C,
  Lawrence ND, Weinberger KQ (eds) Advances in Neural Information Processing
  Systems 27, Curran Associates, Inc., pp 3356--3364,
  \urlprefix\url{http://papers.nips.cc/paper/5596-distributed-bayesian-posterior-sampling-via-moment-sharing.pdf}

\end{thebibliography}


\appendix

\section{Proof of Theorem 1}\label{sc:proof_theorem1}

The claim of the theorem is easily verified by rewriting the equation using the factorizations in Equations~(\ref{eq:stage1})--(\ref{eq:stage3}):
\begin{align*}
&p(\X,\W|\Y) \propto \\
&p\left(\X^{(1)}\right)p\left(\W^{(1)}\right) p\left(\Y^{(1,1)}|\X^{(1)},\W^{(1)}\right) \\
& \times\prod_{i=2}^{I} \left[ p\left(\X^{(i)}\right) p\left(\Y^{(i,1)}|\X^{(i)},\W^{(1)}\right) p\left(\W^{(1)}|\Y^{(1,1)}\right) 
p\left(\W^{(1)}|\Y^{(1,1)}\right)^{-1}\right] \\ 
&  \times\prod_{j=2}^{J} \left[ p\left(\W^{(j)}\right) p\left(\Y^{(1,j)}|\X^{(1)},\W^{(j)}\right)p\left(\X^{(1)}|\Y^{(1,1)}\right)
p\left(\X^{(1)}|\Y^{(1,1)}\right)^{-1}\right] \\
& \times \prod_{i=2}^{I}\prod_{j=2}^{J} \Bigg[p\left(\Y^{(i,j)}|\X^{(i)},\W^{(j)}\right)p\left(\X^{(i)}|\Y^{(1,1)},\Y^{(i,1)}\right) p\left(\W^{(j)}|\Y^{(1,1)},\Y^{(1,j)}\right) \\
&  \phantom{\times \prod_{i=2}^{I}\prod_{j=2}^{J} \Bigg[} 
 \times p\left(\X^{(i)}|\Y^{(1,1)},\Y^{(i,1)}\right)^{-1} p\left(\W^{(j)}|\Y^{(1,1)},\Y^{(1,j)}\right)^{-1}\Bigg] \\
& = \prod_{i=1}^{I} p\left(\X^{(i)}\right) \prod_{j=1}^{J} p\left(\W^{(j)}\right) \prod_{i=1}^{I}\prod_{j=1}^{J} p\left(\Y^{(i,j)}|\X^{(i)},\W^{(j)}\right).
\end{align*}

\section{Hardware platform}\label{sc:compute_node_config}
The experiments for our method and embarrassingly parallel MCMC were performed on Triton (Aalto Science-IT), a cluster with more than 200 compute nodes. The nodes used for the experiments are equipped with dual 12-core Xeon E5 2680 v3, a clock speed 2.50GHz and 128 GB of RAM. One node was used for posterior inference for each subset. The experiments for distributed / parallel implementations of BMF and NMF were conducted on Salomon (IT4Innovations), a cluster with 576 nodes, each equipped with dual 12-core Intel Xeon E5-2680v3, a clock speed 2.50GHz and at least 128 GB of physical memory per node.

\section{Wall-clock time speed-up}\label{sec:run_time_speedup}

\begin{figure}[!htb]
 \begin{center}
 \includegraphics[width=1.\textwidth]{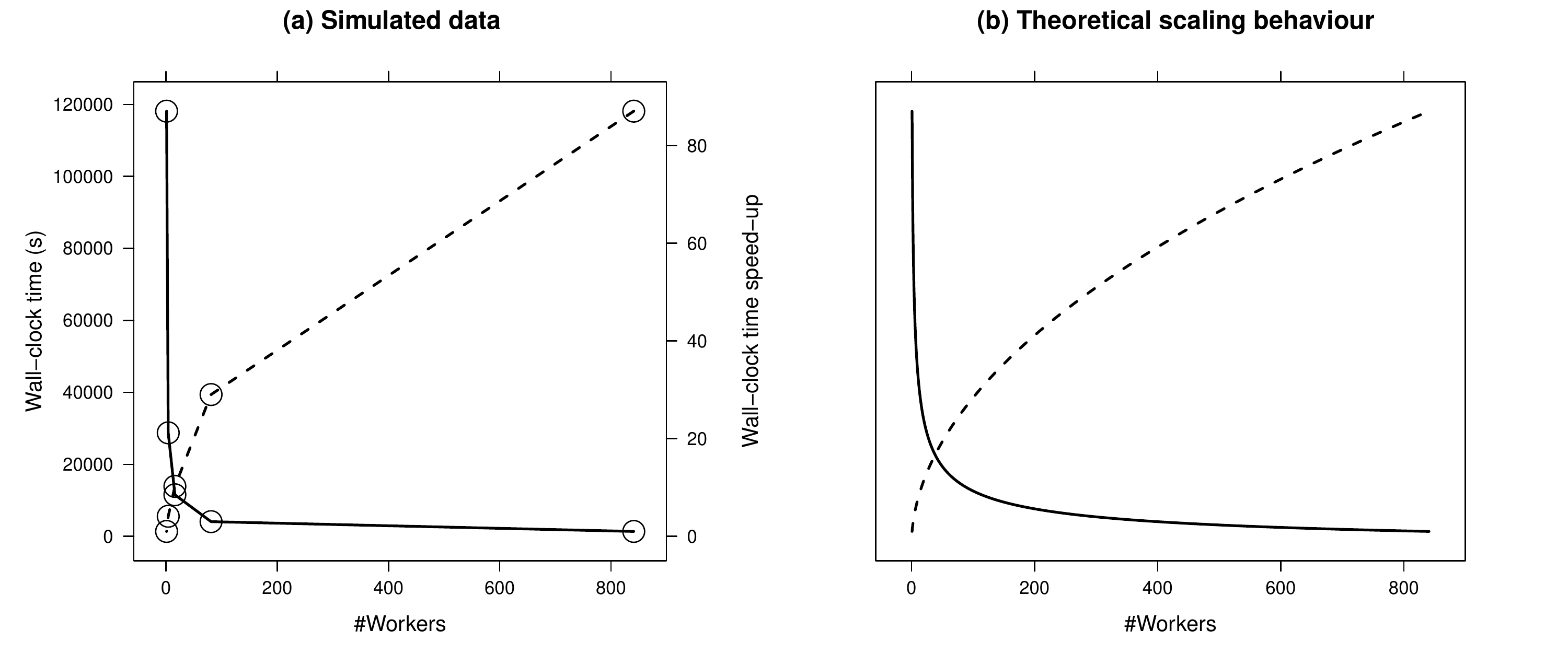}
 \end{center}
 \caption{Scaling behaviour of our method (a) on simulated data with rows and columns in a decreasing order; (b) theoretical scaling behaviour. The x-axis represents the maximum number of parallel workers that can be used by our method for different partition schemes, the left y-axis is the wall-clock time (solid curve), and the right y-axis is the speed-up (dashed curve).}
 \label{fig:wallClockTime_vs_cores_speedup}
\end{figure}

The results for RMSE vs. different partition schemes in Figure \ref{fig:real_data} and \ref{fig:simulated_data} provide a trade-off between predictive performance and computation time. In this section, we revisit the results for simulated data and MovieLens-1M to analyse the scaling behaviour of our method by evaluating wall-clock time speed-up with respect to the number of workers. We follow \citet{Lian:2015:APS:threadSpeedup} and compute the wall-clock time speed-up (WTS) as follows:
\begin{equation}\label{eq:runTime_speedup}
\textrm{WTS} =\frac{\textrm{Wall-clock time of model on the full data}}{\textrm{Wall-clock time of distributed method}} \nonumber
\end{equation}
The results of this analysis are given in Figure \ref{fig:wallClockTime_vs_cores_speedup}(a) and Table \ref{Tab:wallClockTime_vs_cores_speedup}. 
The theoretical scaling behaviour for an arbitrary fixed problem size is shown in Figure~\ref{fig:wallClockTime_vs_cores_speedup}(b), where we use the definition of computation time derived in Section~\ref{sec:scalability} as a function of the number of workers. 
We conclude that the general behaviour in the empirical results is well in line with what we expect from our discussion in Section~\ref{sec:scalability}, with the actual scale factor depending on the problem-specific configurations (e.g. data size, partition scheme), implementation and hardware.

\begin{table}[!htb]
\caption{Wall-clock time speed-up on MovieLens-1M and simulated data.} 
\label{Tab:wallClockTime_vs_cores_speedup}
\renewcommand{\arraystretch}{1.25}
\begin{adjustwidth}{-.5in}{-.5in}
\begin{center}
\begin{tabular}{l|c|c|c|c|c|c|c}
  \hline
\multirow{2}{*}{Partition} & \multirow{2}{*}{\#Workers} & \multicolumn{3}{c}{Decreasing order} & \multicolumn{2}{|c}{Random order} \\ \cline{3-8}
 &  & RMSE & Wall-clock time (s) & WTS & RMSE & Wall-clock time (s) & WTS \\ \hline
\multicolumn{6}{l}{MovieLens-1M} \\ \hline
30x30 & 841 & 0.8881 & 1473 & 23.056  & 0.9194 & 1363 & 24.419 \\ 
10x10 & 81  & 0.8603 & 4529 & 7.498  & 0.8896 & 3710 & 8.971 \\ 
5x5 & 16  & 0.8512 & 10398 & 3.266  & 0.8735 & 8583 & 3.878 \\ 
3x3 & 4  & 0.8485 & 18894 & 1.797  & 0.8605 & 16195 & 2.055 \\ 
1x1 & 1  & 0.8470 & 33956 & 1.0  & 0.8470 & 33283 & 1.0 \\ 
\hline

\multicolumn{6}{l}{Simulated data} \\
\hline
30x30 & 841  & 1.018 & 1357 & 87.069  & 1.020 & 1252 & 94.097 \\ 
10x10 & 81  & 1.009 & 4069 & 29.029  & 1.009 & 4037 & 29.186 \\ 
5x5 & 16  & 1.008 & 11542 & 10.234  & 1.008 & 11467 & 10.276 \\ 
3x3 & 4  & 1.008 & 28749 & 4.109  & 1.008 & 27873 & 4.227 \\ 
1x1 & 1  & 1.008 & 118124 & 1.0  & 1.008 & 117830 & 1.0 \\ 

\hline
\end{tabular}
\end{center}
\end{adjustwidth}
\end{table}

\end{document}